\documentclass[11pt]{article}
\usepackage{CJKutf8}

\usepackage{url}
\usepackage{latexsym}

\usepackage{CJKutf8}
\usepackage{indentfirst}
\usepackage{fancyhdr}
\usepackage{geometry}
\usepackage{booktabs}
\usepackage{multirow}

\newenvironment{chineseabstract}%
  {\vskip 0.125in \centerline{\bf 摘要}\vskip 0.1in
   \leftskip=3em  
   \rightskip=3em 
   }%
  {\vskip 0.3in}

\newenvironment{englishabstract}%
  {\vskip 0.125in \centerline{\bf Abstract}\vskip 0.1in
   \leftskip=3em  
   \rightskip=3em 
   }%
  {\vskip 0.3in}

\newenvironment{keywords}%
  {\noindent\textbf{关键词：}\ }
  {\vskip 0.1in} 

\newenvironment{keywords2}%
  {\noindent\textbf{Keywords}\ }
  {\vskip 0.1in} 

\usepackage{multicol}
\usepackage{array}
\usepackage{longtable}

\usepackage[table]{xcolor}
\usepackage{booktabs}
\usepackage[english]{babel}

\usepackage{graphicx}
\usepackage{epstopdf}
\usepackage{hyperref}

\addto\captionsenglish{

    }

\title{大语言模型训练数据}

\author{鞠一鸣， 马欢欢 \\ \\ yiming.ju2024@gmail.com}

\date{2024.11.12}

\begin{document}

\begin{CJK*}{UTF8}{gbsn}
\setlength{\parindent}{2em}

\maketitle

\begin{chineseabstract}
2022年，随着ChatGPT的发布，大规模语言模型受到了广泛关注。ChatGPT不仅在参数量和预训练语料库规模上远超前代模型，还通过大量高质量的人工标注数据进行微调，实现了革命性的性能突破。此类进展使得企业和研究机构认识到，构建更智能、更强大的模型依赖于丰富且高质量的数据集。因此，数据集的建设与优化成为人工智能领域的关键方向。本文对训练大规模语言模型所需的预训练数据和微调数据的现状进行了总结，涵盖了数据规模、搜集方式、数据类型及特点、处理流程等，并对当前可用的开源数据集进行了梳理和介绍。


\end{chineseabstract}

\begin{englishabstract}
In 2022, with the release of ChatGPT, large-scale language models gained widespread attention. ChatGPT not only surpassed previous models in terms of parameters and the scale of its pretraining corpus but also achieved revolutionary performance improvements through fine-tuning on a vast amount of high-quality, human-annotated data. This progress has led enterprises and research institutions to recognize that building smarter and more powerful models relies on rich and high-quality datasets. Consequently, the construction and optimization of datasets have become a critical focus in the field of artificial intelligence. This paper summarizes the current state of pretraining and fine-tuning data for training large-scale language models, covering aspects such as data scale, collection methods, data types and characteristics, processing workflows, and provides an overview of available open-source datasets.

\begin{keywords2}
Language model, Training data, Pre-training, Fine-tuning
\end{keywords2}
  
\end{englishabstract}

\tableofcontents
\newpage

\section{引言}

在人工智能领域，语言模型的发展一直是推动技术进步的核心动力之一。语言模型的目标是通过模拟文本数据的生成概率，实现对自然语言的理解和生成。最初的模型，如n-gram，依赖于统计分析来预测词语序列，但这些模型难以捕捉复杂的文本依赖性，因此在处理复杂任务时表现不佳。
随着深度学习技术的兴起，特别是
AlexNet
在视觉识别领域的突破，神经网络被引入到自然语言处理中。循环神经网络（RNN）及其衍生的LSTM和GRU显著提升了模型处理长序列数据的能力，使其能够更好地建模语言中的时序特征和文本关系。然而，这些方法仍然依赖于传统的监督学习范式，训练数据规模普遍较小，大部分数据集的规模以MB为量级。

真正的革命来自于以Transformer\cite{vaswani2017attention}为基础的预训练语言模型的推出。Transformer架构通过自注意力机制（Self-Attention）有效地捕捉长距离依赖关系，并且由于其高度并行化特性，极大地提高了语言模型的训练效率。这使得以大规模无标记文本作为训练语料成为可能。GPT和BERT\cite{kenton2019bert}等模型以Transformer为基础，大幅提升了语言模型的性能，将语言模型的“智力”推向了一个新的高度。
预训练语言模型之所以强大，是因为它们能够在未经人工标注的大数据集中自动学习丰富的特征。这些模型在预训练阶段通过处理大量数据发现内在规律，培养出广泛的应用能力和强大的表征能力。然而，预训练模型虽然蕴含了丰富的语言知识，但并不能直接用于解决问题或完成特定任务。因此，需要在特定领域的数据集上进行微调，使模型能够更好地适应特定应用需求。

2022年，随着ChatGPT\cite{openai2022introducing}的发布，大规模语言模型开始受到广泛关注。ChatGPT不仅拥有远超之前模型的参数量，其使用的预训练语料库规模也更为庞大，并且使用大量高质量的人工标注语料进行微调。ChatGPT革命性的性能突破让企业和研究机构认识到，构建更智能、更强大的模型需要依赖于更丰富、更高质量的数据集。因此，数据集的建设和优化成为了人工智能领域的重要发展方向。
在这一背景下，许多公司和组织参与到语料数据的构建和优化工作中。虽然很多大规模语言模型（如Baichuan\cite{bib5}、Qwen\cite{bai2023qwentechnicalreport}、Chatglm\cite{glm2024chatglmfamilylargelanguage}等）已经相继开源，但其训练数据几乎全部闭源并视其为重要的商业机密。
可以说，在统一使用Transformer作为模型架构的背景下，高质量的训练数据是开发大语言模型和保持业内领先的核心竞争力。
当然，也有非常多的公司和组织把投入大量资源构建的数据进行开源，为社区的发展做出贡献，
如Together AI\cite{togetherai}和Hugging Face等公司基于Common Crawl\cite{commoncrawl}项目对数据集进行清洗和优化，并发布了RedPajama\cite{bib16}数据集和Hugging Face的FineFeb\cite{penedo2024finewebdatasetsdecantingweb}数据集。这些数据集不仅去除了噪声和不相关内容，还进行了结构化和标准化处理，使其达到了预训练所需的数据质量标准。

虽然目前国内外已经有非常多的预训练语料数据与微调语料数据，也有一种声音称全世界的未被开发语料数据已经所剩无几。然而，数据集的建设，维护和优化是一个持续的过程。随着大模型技术的发展，对新数据的需求也在不断增加。虽然目前的语料数据规模已经非常庞大，未来的数据集建设可能更加注重数据分布多样性、数据质量和可解释性。同时，随着数据安全和隐私保护的重要性日益突出，如何在保护用户隐私的前提下合理利用数据资源，也成为了数据集建设中的一个重要考虑因素。

\section{预训练数据}

预训练任务通过让模型基于已有文本预测未知的Token，以此来学习语言结构和语义规则。自编码训练任务（如BERT）和自回归训练任务（如GPT）分别代表了两种主要的方法。在ChatGPT发布后，自回归训练任务逐渐成为训练大规模语言模型的主流方法。
在预训练过程中，模型通过执行预测Token的原子任务，逐步掌握文本的语法和语义，从而积累大量语言知识，提升文本理解和生成能力。核心机制在于，模型能够通过预测文本的一部分，逐渐建立对整个文本结构的理解。借此，模型不仅能记住具体的单词和短语，还能学习更高层次的语言结构和语义关系。这种能力对于生成连贯且有意义的文本至关重要，使得模型在实际应用中能够提供逻辑合理、上下文一致的回答。

预训练数据对于训练模型理解和生成自然语言方面的能力至关重要，是大规模语言模型开发的基础。预训练语料数据未经人工标记，含有丰富的语言元素和复杂的结构。这样的非标记数据为模型提供了真实、自然的语言使用场景，使其能够学习到语言的本质特征和潜在规则。
预训练语料的首要特征就是规模庞大。大规模的数据集能够提供丰富的语言现象和多样的上下文场景，使模型在训练过程中接触到各种语言结构、词汇用法和语法规则。这样的规模保证了模型可以捕捉到更广泛的语言模式和关系，从而在多种任务中表现出色。
除了规模庞大，其第二个重要特征就是来源广泛。预训练语料涵盖了从日常通信到专业学术的各种文本类型，这包括新闻报道、社交媒体帖子、科学论文、文学作品等。这种多样化的文本来源确保了模型能够在不同的语言环境中灵活应用，既能理解口语化的表达，也能处理专业术语和复杂的学术论述。模型通过在这些不同类型的文本中学习，可以获得更全面的语言知识和更强的适应能力。

通过预训练，模型能够积累大量的背景知识，这使得它在面对新的任务时，能够迅速适应并表现出色。预训练就像是让模型读了成千上万本书，见多识广，积累了丰富的知识储备，而微调过程则是让模型在特定领域进行专门训练，以便更好地完成特定任务。
此外，预训练数据的多样性和复杂性也意味着模型在学习过程中需要处理一些含有噪音或者不规则的数据。尽管如此，其中一些噪音可能是模型成长的重要部分，如错别字，模糊的表达等。通过在复杂、多变的语言环境中学习，模型能够变得更加健壮和灵活，能够更好地处理实际应用中遇到的各种问题。


\subsection{语料类型与来源}

预训练语料库由来自不同领域和来源的大规模文本数据组成。这些数据集的主要特点是数量庞大，并且分布广泛，不局限于某一领域，使其适合用于训练通用的基础模型。通常来说，预训练数据类型可以分以下几个主要类别：
\begin{itemize}
\item网页:包含来自互联网的丰富信息，覆盖新闻、博客、论坛帖子等多种网络文本形式，反映了当代社会的多元视角和实时动态。
\item书籍：囊括了小说、教科书等各类图书，它们不仅提供了深厚的文化和知识底蕴，还涵盖了广泛的主题和复杂的叙事结构。
\item学术资料：涵盖了学术论文、期刊文章、会议论文等，这些资料代表了特定领域的前沿知识和研究成果。
\item代码：包括各种编程语言的源代码和文档，为模型提供了理解和生成技术性文本的能力。
\item 社交媒体：集成了来自Twitter\footnote{Twitter是一个社交媒体平台，用户可以在其中发布短消息，网址为：\url{https://twitter.com/}}、Facebook\footnote{Facebook是一个全球最大的社交网络平台之一，网址为：\url{https://www.facebook.com/}}、微博\footnote{微博是一个中国的社交媒体平台，网址为：\url{https://weibo.com/}}等社交平台的用户生成内容，反映了人们的日常生活和即时反应。
\item 百科：如维基百科\footnote{维基百科是一个开放的、自由编辑的百科全书项目，网址为：\url{https://www.wikipedia.org/}}、百度百科\footnote{百度百科是一个中文在线百科全书，网址为：\url{https://baike.baidu.com/}}等，提供了结构化的知识条目，覆盖了广泛的主题和定义。

\end{itemize}


下面分别介绍这八种类型的预训练数据。
\subsubsection{网页数据}

网页数据来自互联网抓取的大量网页，是预训练语料库中最常见和最广泛使用的数据类型。网页数据能够作为预训练语料库最主要的数据类型有以下几个原因：

\begin{itemize}
\item规模庞大：互联网经过几十年的发展，已经存在数以亿计的网站，使得网页数据规模极其庞大。
\item主题广泛：互联网几乎已经深入到人类生活中的方方面面，使得网页内容覆盖了从日常生活、娱乐到科学、技术、医学等各个领域。
\item用户生成内容：互联网充满了用户生成的内容，如社交媒体帖子、评论、博客文章和论坛讨论，反映了真实的语言使用和社会互动模式。
\item多语言性：
互联网是一个多语言的平台，网页数据包含了世界各地的多种语言内容，为训练多语言模型提供了基础。
\item动态更新：每天都有新的网页产生和消失，这种动态性使预训练模型能够接触到最新的信息和语言表达方式。
\item数据公开性：很多网页数据可以通过合法的方式进行抓取和使用，具有较高的公开性和可获取性。

\end{itemize}

目前网页语料库构建的主要数据来源是Common Crawl\footnote{Common Crawl是一个非盈利组织，网址为：\url{https://commoncrawl.org/}}。Common Crawl是一个非盈利的组织，致力于定期抓取和存档全球互联网上的公开网页数据。这个项目始于2007年，该数据集包括网页的原始HTML、文本内容、元数据以及各种其他信息。以下是有关Common Crawl的一些主要特点：

\begin{itemize}

\item数据量庞大：Common Crawl每个月都会进行新的抓取，每月抓取的数据量达到数十TB。

\item数据公开可访问：任何人都可以免费下载并使用Common Crawl数据。

\item格式统一：Common Crawl数据通常以WARC（Web ARChive）文件格式存储，包含抓取的网页内容和相关的元数据。

\item中文占比少：中文网页占比只有大约 1.4\%，远低于英文网页的 55\%。

\end{itemize}

Common Crawl作为一个开放的网页爬取数据仓库，为用户提供了一个宝贵的资源，使得大规模的网页内容分析成为可能。通过提供这些数据，Common Crawl支持了各种研究和开发活动，包括但不限于搜索引擎优化、语言模型训练、社会数据研究等。由于其数据量巨大且多样化，Common Crawl成为了许多数据驱动项目的重要基础，在GPT-3的训练中，Common Crawl贡献了其中约60\%数据的数据来源。
然而，Common Crawl中的数据并不能直接用于语言模型的训练，因为其致力于抓取所有网页的一个尽可能大的子集，并不会对网页内容进行过滤。其原始数据包含大量不适合作为训练预料的内容，如广告、导航栏等无关信息；色情内容、暴力等有害信息；机器生成的垃圾邮件和广告等无价值内容；涉及个人隐私的敏感信息等。
目前，很多预训练语料库通过过滤和清洗Common Crawl的数据构建。例如，RefinedWeb数据集对Common Crawl进行了严格的过滤和去重处理，最终保留了总计5T tokens的高质量英文文本用于预训练Falcon模型；
RedPajama-1T数据集中的网页数据同样来源于Common Crawl，其使用CC-NET框架处理了Common Crawl的五个抓取结果，主要包括文本提取、去重、语种分类、过滤低质量语料以及结构化输出等步骤，最终获得了0.88T tokens的英文文本。
Pile-CC使用训练的分类器来筛选低质量文本，从而实现对Common Crawl内容的过滤。













\subsubsection{书籍}

书籍数据在预训练语料库中扮演着重要角色。相较于网页内容，书籍提供了更长的文本和更高的质量，这对于提升大型语言模型的性能至关重要。书籍数据能够帮助模型更准确地捕捉语言特征，并深入学习丰富的语言知识和上下文信息。以下是书籍数据的一些主要特点：

\begin{itemize}

\item主题广泛：书籍通常涵盖广泛的主题和领域，包括小说、传记、教科书等，提供了多样化的内容。

\item文本质量高：由于书籍通常由专业作者撰写，并经过编辑和校对，因此其文本质量更高，语法和拼写错误较少，噪音较少。

\item文本长度长：书籍中的文本长度较长，句子结构更复杂，能够提供更多的上下文信息，有助于模型理解和生成复杂的语言结构和建模远距离的依赖关系。

\item多样的语言和文化特征：书籍中包含丰富的语言特征，如专业术语、俚语和成语，反映了不同的文化背景，增加了语料库的多样性。

\item知识性强：书籍数据包括人类各个领域的教材，工具书，百科等。有着很强的知识性，有助于帮助模型学习各个领域的知识。

\end{itemize}

目前书籍数据通常通过处理电子书数据获得，电子书数据有着多种获取途径，下面介绍几个著名的电子书资料库：

\begin{itemize}
\item
Smashwords\footnote{Smashwords是一个电子书出版和分发平台，允许作者自助出版和销售电子书，网址为：\url{https://www.smashwords.com/}。}
是一个大型的电子书平台，提供了超过50万本电子书，涵盖各种主题和类型。该平台允许独立作者和出版商自助发布电子书，从而为读者提供丰富的阅读选择。Smashwords以其开放性和易用性吸引了众多作者和读者，是获取多样化书籍数据的宝贵资源。

\item Project Gutenberg\footnote{Project Gutenberg是一个提供免费电子书的数字图书馆，包含大量版权过期的经典文学作品，网址为：\url{https://www.gutenberg.org/}。}
是世界上最早的数字图书馆之一，致力于数字化和存档公共领域的文化作品。该项目由志愿者运营，提供了超过6万本免费电子书，涵盖经典文学、历史文献、科学著作等。Project Gutenberg的目标是促进文化和知识的传播，为读者和研究人员提供方便的访问途径。

\item Z-Library\footnote{Z-Library是一个数字图书馆，提供大量免费和付费的学术、技术和文学书籍，网址为：\url{https://z-lib.id/}。}
是一个在线电子书库，号称拥有世界上最大的电子书和文章集合。该平台提供了数百万本书籍和学术论文，涵盖广泛的主题和学科。尽管Z-Library因其内容广泛和免费访问的特点受到欢迎，但其合法性和版权问题也引发了一些争议。

\item Sci-Hub\footnote{Sci-Hub是一个提供免费学术论文访问的平台，旨在打破学术出版物的收费壁垒，网址为：\url{https://sci-hub.se/}。}
是一个提供学术论文和研究文章免费下载的平台，由Alexandra Elbakyan于2011年创建。Sci-Hub旨在打破学术出版的付费墙，使科研成果可以免费获取。尽管Sci-Hub在学术界受到欢迎，帮助了许多研究人员，但也因其侵犯版权的行为而面临法律挑战。

\item创世纪图书馆\footnote{创世纪图书馆（Anna's Archive）是一个数字图书馆项目，致力于存档和提供免费电子书，网址为：\url{https://annas-archive.org/}。}
，又称Anna’s Archive，是一个开源和开放数据的数字图书馆。它是世界上最大的开放数据图书馆之一，创建者从多个平台（如Libgen、Sci-Hub）抓取书籍并公开提供。截至2024年2月，Anna’s Archive的规模已达到641.2TB，并且还在不断增长。该平台致力于保存和传播文化和知识，为全球读者提供丰富的资源。

\item
PG-19 是一个大型的历史书籍语料库。收录的书籍全部出版于1919年之前，这些书籍涵盖了广泛的历史时期和主题，包括文学、科学、历史、哲学等多个领域。这样的语料库为研究语言演变和历史文本分析提供了宝贵的数据。

\end{itemize}

这些平台为预训练语料库提供了丰富且多样的书籍和学术资源。它们在数据质量、主题广度和可访问性方面各具特色，为研究人员和开发者提供了宝贵的数据来源。同时也需要注意版权和合法性问题，在使用这些资源时应遵循相关法律法规。
相比于网页数据，书籍数据的噪声相对较少。书籍数据的后处理主要包括去除封面、目录、版权页等元信息。此外，一些电子书资源可能会在书籍中添加广告，需要进行有针对性的过滤。
由于书籍数据的分布并不均衡，例如，中文电子书中包含一些长期连载的网络小说，这些书籍的长度远超其他中文书籍。在使用书籍数据前，需要对数据进行调整和筛选，以确保数据的平衡性。
\begin{table}[ht]
\centering
\caption{中国图书分类法}
\small
\begin{tabular}{>{\bfseries}c>{\raggedright\arraybackslash}p{0.8\textwidth}}
\toprule分类代码 & 分类名称 \\
\midrule
A & 马克思主义、列宁主义、毛泽东思想、邓小平理论 \\
B & 哲学、宗教 \\
C & 社会科学总论 \\
D & 政治、法律 \\
E & 军事 \\
F & 经济 \\
G & 文化科学、教育、体育 \\
H & 语言、文字 \\
I & 文学 \\
J & 艺术 \\
K & 历史、地理 \\
N & 自然科学总论 \\
O & 数理科学和化学 \\
P & 天文学、地球科学 \\
Q & 生物科学 \\
R & 医药、卫生 \\
S & 农业科学 \\
T & 工业技术 \\
U & 交通运输 \\
V & 航空、航天 \\
X & 环境科学、安全科学 \\
Z & 综合性图书 \\
\bottomrule
\end{tabular}
\label{book_cls}
\end{table}

书籍内容广泛，并且已有较为完备的分类体系。如表 \ref{book_cls}，根据中国图书馆分类法，书籍可以分为“语言、文字”、“哲学、宗教”、“文学”等22个类别。其中，每一个类别还有更详细的子类别，如“历史、地理”类别包含史学理论、世界史、中国史、亚洲史等。按领域进行细致分类有助于更好地利用数据。虽然并不是所有书籍数据都带有详细的类别标签，但可以使用已有标签的数据训练分类模型，然后对其他数据进行类别分类进而获得所需要的标签。
在构建数据集时，因为小说类文本长度较长占比较多，常见的方法是降低小说类书籍的比例，增加其他书籍的比例。这种调整可以确保数据集更具多样性和实用性，有助于提升预训练模型的性能。

\subsubsection{学术资料}

学术资料数据也是预训练数据的重要组成部分，其包含学术论文、期刊文章、会议论文、研究报告、专利等各种文本数据。这些数据由学术界的专家和学者撰写和发布，具有高度的专业性和学术严谨性。由于学术资料本身的高质量和专业深度，将其纳入预训练语料库中可以为模型提供更加准确和专业的信息，帮助模型更好地理解和掌握学术领域的术语和知识。学术资料数据有着以下特点：

\begin{itemize}
\item
\textbf{高质量}：学术资料由领域内的专家撰写，经过同行评审和编辑，具有非常高的文本质量和内容价值。即使是未经评审的学术资料，其语法和结构也十分规范。这种高质量的数据有助于提高预训练模型的准确性和可靠性。

\item
\textbf{专业性}：学术材料涵盖各个学术领域的最新研究成果和前沿进展，包含大量的专业术语和学术表达。通过学习这些数据，模型可以掌握更多专业知识，从而提高其在特定领域任务中的表现。

\item
\textbf{丰富的图表和公式}：学术材料通常包含大量的图表和对其的详细分析，这可以帮助模型理解表格和其他结构化文本，提高其数据处理能力。

\item
\textbf{特殊格式}：许多学术材料以LaTeX等格式储存，这有助于模型理解和生成LaTeX格式的文本，提升其对特殊文本书写方式的掌握。
\end{itemize}

目前构造学术资料数据最常用的语料来源为arXiv\footnote{arXiv: https://arxiv.org}。
arXiv是一个开放获取的预印本存储库，最初于1991年由物理学家Paul Ginsparg创建，主要用于物理学领域的论文共享。随着时间的推移，arXiv的覆盖范围扩展到了数学、计算机科学、生物学、统计学、金融数学和其他定量领域。arXiv的主要目的是促进科学研究的传播和交流，为研究人员提供一个快速发布和获取最新研究成果的平台。
arXiv能够成为预训练语料库中学术资料数据的主要数据来源的原因主要有以下三点：

\begin{itemize}
\item
arXiv的开放获取政策允许任何人免费访问和下载其全部内容并用于构建训练语料。此外arXiv还提供了便捷的API和批量下载工具。

\item
arXiv上的很多论文使用LaTeX格式编写，研究人员可以直接获取LaTeX源文件。

\item
arXiv上已经收录了超过200万篇论文，有着庞大的数据量且还在不断增长。
\end{itemize}

需要注意的是，不同于常见的期刊和会议语料库，研究者可以相对自由地上传其研究成果到arXiv，有一些语料并没有经过严格的内容审查。
除arXiv外，也有一些学术资料语料库可以被用来构建预训练语料：

\begin{itemize}
\item
S2ORC是一个包含来自各个学科的英文学术论文的大型语料库。由Semantic Scholar团队创建。这个语料库的一个显著特点是其广泛的覆盖范围和高质量的结构化数据\footnote{S2ORC: https://github.com/allenai/s2orc}.
\item
PubMed Central （PMC） 是一个免费的生物医学和生命科学期刊文献存档，由美国国立卫生研究院 （NIH） 管理。PMC包含了大量的开放获取期刊文章，是生物医学研究的重要资源\footnote{PubMed Central: https://www.ncbi.nlm.nih.gov/pmc}.
\item
IEEE Xplore是一个大型的技术文献数据库，由电气电子工程师协会 （IEEE） 管理。该数据库包含了大量的工程学、计算机科学和电子技术领域的文献\footnote{IEEE Xplore: https://ieeexplore.ieee.org}.
\end{itemize}

需要注意的是，使用这些学术资料语料库时需要注意其版权和使用许可范围。此外，学术材料数据的分布可能不均衡，某些领域的文献数量可能远超其他领域。例如，尽管arXiv涵盖了不同研究领域的学术资料，其收录论文相对集中在物理学、数学和计算机科学等领域。
在使用这些数据之前，需要对数据进行调整和筛选，确保预训练语料库的平衡性和多样性。

\subsubsection{代码}
代码数据指的是包含编程语言的文本信息，如Python、Java、C++等代码片段。这些数据有助于模型更好地理解编程语言和代码结构，从而在代码理解、代码推荐和代码生成等下游任务中表现出色。随着大规模语言模型的广泛应用，代码生成已成为其重要功能之一，训练模型时使用的代码数据质量直接影响生成代码的效果。因此，代码数据在预训练数据中具有重要地位。此外，有研究推测，大语言模型出色的逻辑能力与其在大量的代码数据上进行了训练有关。

\begin{itemize}
\item The Stack\cite{bib31}是一个多样化的数据集，包含多达385种编程语言的源代码文件，总量超过6TB。所有代码文件都具有开源许可证。该数据集涵盖了各种编程语言，从流行的Python和Java到更专业的语言如Fortran和Lisp。这种多样性有助于模型学习广泛的编程语法和风格。

\item Github是一个托管开源和私有软件项目的平台，提供了丰富且多样的项目。使用Github里的代码数据时需要遵循项目作者设定的使用协议。目前已有许多基于Github构造的代码训练语料：Google BigQuery\cite{google_bigquery_github}提供了大量已通过SQL查询进行过滤的GitHub许可存储库的快照；CodeParrot\cite{codeparrot}包含包含数百万个来自于GitHub的代码片段；CodeSearchNet\cite{husain2020codesearchnetchallengeevaluatingstate}是一个关注代码搜索任务的数据集，也基于Github创建。

\item StackOverflow\cite{stackoverflow}是一个专注于编程和开发的问答社区，涵盖了编程语言、开发工具和算法等话题。该平台不仅是代码片段的丰富来源，还包含了大量的开发知识和经验分享。由于其社区性质，StackOverflow的数据也可以被归类为社交媒体数据的一部分。

\item ProjectCodeNet\cite{ibm_project_codenet}是一个大规模、多语言的代码数据集，旨在促进代码理解和生成任务的研究。它由IBM提供，涵盖广泛的编程语言和多种代码任务。

\end{itemize}

如今，代码生成能力已经是大语言模型最重要的能力之一，在提高相关人员工作效率上起着重要作用。
代码数据在大规模语言模型的训练中具有不可替代的重要性。通过使用The Stack、StackOverflow等丰富的代码相关语料，模型能够学习和理解不同编程语言的结构和语法。这不仅提高了模型在代码相关任务中的表现，同时也可以训练模型的逻辑推理能力。

\subsubsection{社交媒体}

社交媒体数据指的是从各种媒体平台收集的文本内容，主要包括用户生成的帖子、评论以及用户之间的对话数据。这些数据反映了社交媒体上的实时动态和互动情况。尽管社交媒体数据中可能存在偏见、歧视和暴力等有害信息，但它仍然是大规模语言模型（LLMs）预训练的重要资源。这是因为社交媒体数据有助于模型学习人类在各种场景下的表达方式，提高模型的对话生成能力、捕捉社交趋势和用户行为模式等。社交媒体数据有着以下特点：

\begin{itemize}
\item \textbf{多样性}：
社交媒体上的用户生成内容涵盖日常生活到专业知识等广泛的主题，并且包含不同用户在不同场景下的自由表达，有着丰富的语言风格和表达形式。例如，社交媒体数据包含很多俚语、网络用语、缩写和新造词。这些文本有助于增加预训练语料库文本风格的多样性。

\item \textbf{互动性}：
社交媒体平台上用户之间的对话数据提供了丰富的互动场景，涵盖问答、讨论、辩论等多种形式。通过学习这些对话数据，模型可以更好地理解人类的交流方式。例如，模型可以学会如何在处理多轮对话，理解上下文，从而生成更加连贯和有意义的对话内容。

\item \textbf{噪声多}：
由于社交媒体平台上的内容未经严格筛选和编辑，很可能包含拼写错误、语法错误和不相关的信息。这些噪声数据在一定程度上增加了数据处理的难度，但也为模型提供了学习处理和忽略噪声的训练场景。
\end{itemize}

以下是几个常见的社交媒体语料库：
\begin{itemize}
\item知乎是一个高质量的中文问答平台，在这个平台上，用户可以提出问题、回答问题，并分享他们的见解和知识。知乎的内容涵盖了广泛的主题，其内容质量较高，常常伴随着详细的解释和专业的见解。

\item微博是我国最流行的社交媒体平台之一，用户可以通过短文本发布内容，其涵盖时事新闻、娱乐八卦、生活动态等各种信息。。

\item StackExchange是一个包含各种领域和主题的一系列问答平台，是互联网上最大的公开问答数据集之一。专注代码问题的StackOverflow就是其子平台之一。

\item Reddit是一个包含大量用户生成帖子的社交平台。用户可以在各种主题的子论坛（subreddits）中发表帖子和评论，进行讨论和分享信息。

\end{itemize}

除了上述平台，国内外还有许多社交媒体平台拥有大量的语料，例如百度贴吧、小红书、大众点评、Facebook和Twitter等。然而，这些社交平台上的语料并非全部公开可用，在使用时需要注意其许可范围。此外，即使是这些社交媒体平台的运营公司在使用其内部数据时，也存在着一些争议：使用这些平台上的数据是否需要获得平台用户的同意？这些数据是否涉及平台用户的隐私？例如，Meta作为Facebook的母公司，因在英国滥用数据而被罚款13亿美元。此外，Sora是否使用了YouTube和Instagram上的数据也引发了很大争议。这些现象表明，随着领域的发展，数据隐私问题正受到越来越多的关注。由于各国的法律法规不同，如何合规地使用社交媒体数据也是一个值得注意的重要事项。

\subsubsection{百科}

百科数据指的是从百科全书和在线百科网站提取的文本信息。在线百科网站的数据由专家、志愿者和社区贡献者共同编写和编辑，具备一定的权威性和可靠性并且获取方便。这些数据是预训练语料库重要组成部分。百科数据具有以下特点：

\begin{itemize}
\item知识性：
百科数据通常包含了大量详细的知识。这些知识涵盖广泛的领域，如自然科学、人文艺术等。

\item权威性和可靠性：
百科数据通常由专家、志愿者或社区贡献者编写和编辑，具备一定的权威性和可靠性。

\item易获取性：
在线百科网站的数据可以方便地获取，其内容通常是免费和公开的。

\item多语言支持：
Wikipedia提供多种语言版本，能够为多语言模型的训练提供丰富的语料支持。

\item结构化：
百科数据通常有较为规范组织层次结构，条目内容清晰，引用和参考文献完整。

\end{itemize}

目前百科数据主要来源于百度百科和Wikipedia这两个在线百科网站。
百度百科是百度公司推出的一部内容开放、自由的网络百科全书。截至2023年4月，百度百科已经收录了超2700万个词条，参与词条编辑的网友超过770万人，几乎涵盖了所有已知的知识领域，是中文领域最为广泛使用的在线百科全书。
Wikipedia是全球范围内使用最广泛的百科资源之一，以其免费、开源、多语言和高文本价值而著称。通常，研究人员会选择特定语言的Wikipedia数据，通过爬取和过滤的方式，将其纳入预训练语料库。百科数据在大规模语言模型的预训练中扮演了重要角色，不仅提供了权威可靠的知识来源，还因其易获取性而被广泛应用。这些特性使得百科数据成为增强语言模型知识基础的宝贵资源。

\subsubsection{其他数据}
除以上6种常见类别外，也有一些其他语料可以被纳入大语言模型的预训练范围内，其使用与否取决于对模型特性的考量。
\begin{itemize}
\item平行语料库：平行语料指彼此互为翻译的不同语言的文本对。ParaCrawl\cite{espla-etal-2019-paracrawl}利用开源软件抓取网页，构建了包含2.23亿句子对的公开平行语料库；
MTP\cite{baai_mtp}则收集和整理现有的中英网页文本数据，汇集了3亿个文本对，是目前最大的开源中英对齐文本数据集；
MultiUN\cite{eisele-chen-2010-multiun}通过联合国官方文件系统收集平行文本对；UNCorpus v1.0\cite{ziemski-etal-2016-united}则由联合国官方记录和其他会议文件组成。

\item领域特定语料库：
领域特定语料库指某一特定领域的专有语料，如金融，法律，医学等。
金融领域语料多为金融新闻、财务报表、公司年报、金融研究报告、金融文献、市场数据等，有BBT-FinCorpus\cite{bib43}，XuanYuan\cite{bib44}，FinGLM\cite{bib26}，TigerBot\cite{chen2023tigerbotopenmultilingualmultitask}等语料资源。
医疗语料通常来源于医学文献、健康诊断记录、病例报告等，有Medical-pt\cite{bib45}，PubMed Central\cite{pubmed_central_pmc}等语料资源。
法律语料通常来源于法律文件、法律书籍、法律条款、法院判决和案例等。
\end{itemize}

平行语料库主要用于增强模型的多语言能力和翻译能力，而领域特定语料库通常用于训练专注于特定领域的语言模型。
如果需要将模型应用于特定领域的下游任务，可以在通用预训练语料库的基础上增加领域预训练语料库进行增量训练。通过引入大量领域相关的文本，模型能够更好地理解该领域的专有术语、概念和背景知识，还能提升模型生成领域相关文本的专业性和流畅度，从而提高处理领域特定任务的能力。
例如，在医疗领域，通过使用包含医学术语和案例的语料库，模型能够更准确地理解医学专有名词并生成更为流畅和更符合医学领域规范的文本。



\subsection{预训练数据处理}
从不同的来源收集到所需语料数据后，需要进行数据处理以提高数据质量，同时减少有害内容和敏感内容。此外，还可以给每一条数据增加数据质量，数据类别等详细标签，以便于对数据进行筛选和分析。本节介绍一些较为常见的预训练数据处理方法。

\subsubsection{网页数据处理}
\begin{figure}[htb]
\centering
\includegraphics[scale=0.6]{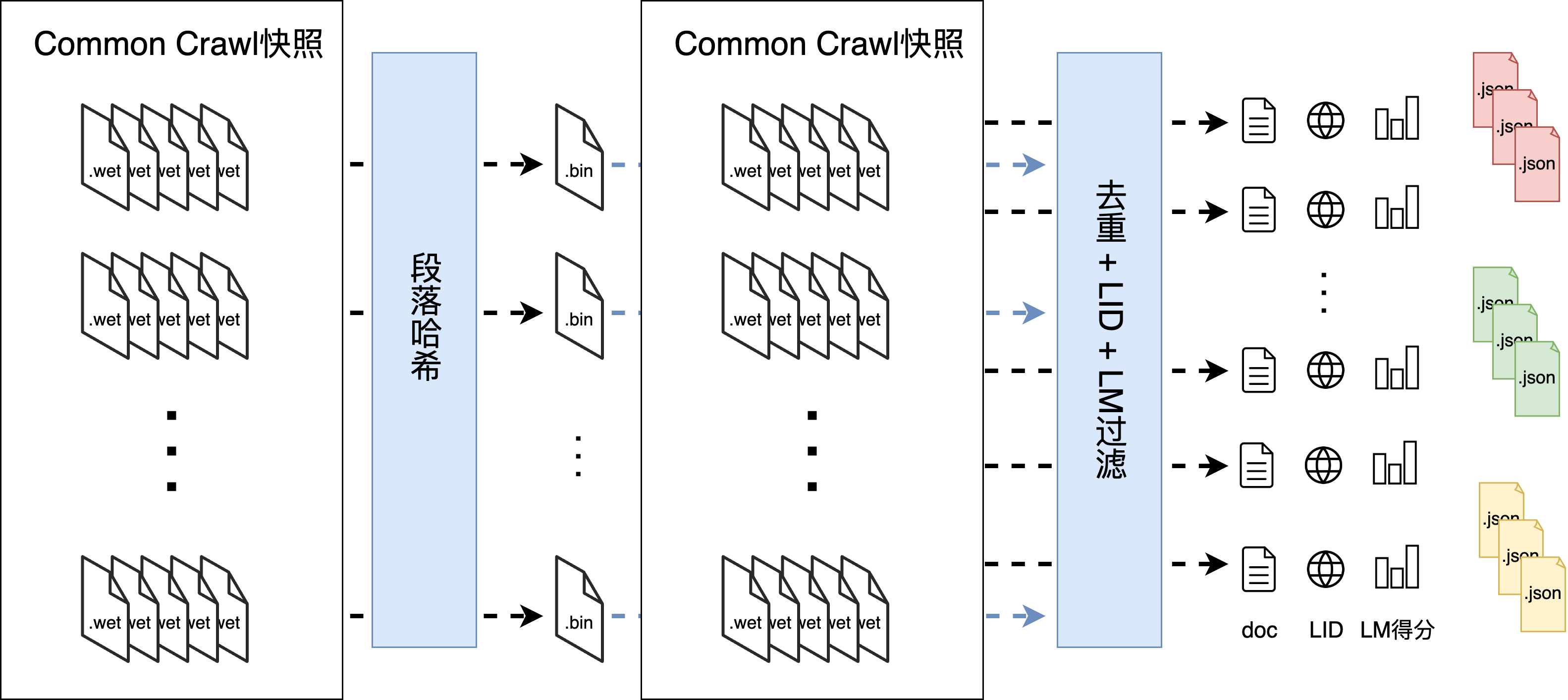}
\caption{CCNet\cite{wenzek-etal-2020-ccnet}处理一个Common Crawl快照的流程}
\label{CCNet}
\end{figure}

网页数据作为大语言模型预训练语料最为重要的数据来源，其数据量也是最大的，并且有着最多的噪声。网页数据通常通过处理Common Crawl的多个快照得到，如RedPajama-V2\cite{bib16}处理了84个Common Crawl快照，每个Common Crawl快照的文本大小约300T。处理并清洗如此巨大的数据是一个复杂的任务，同时也是构建数据集的核心任务之一，对确保数据质量和提高语言模型的训练效果至关重要。
语料处理的主要步骤包括文本提取、语种分类、去重和过滤等。目前，开源语料清洗框架有SlimPajama\cite{shen2024slimpajamadcunderstandingdatacombinations}、MNBVC\cite{bib23}和 CCNET\cite{wenzek-etal-2020-ccnet}等。
本小节首先介绍一个最为著名的处理网页数据的处理框架: CCNet。很多数据集的构建都遵循该pipeline的方法处理网页数据，包括处理多来源的全量数据，如LLama\cite{touvron2023llama}，RedPajama\cite{bib16}等。

CCNet最早介绍于FaceBook 2019年的工作 “CCNet: Extracting High Quality Monolingual Datasets from Web Crawl”，用于处理Common Crawl网页数据。
Common Crawl提供了多种可供使用的数据集格式。目前，主要有三种不同的主要格式：
\begin{itemize}
\item WARC: 用于存储和传输Web资源（例如HTML页面、图像和视频文件等）的文件格式，通常包含HTTP响应和元数据，记录了由Web爬虫收集的信息。
\item WAT: 一种元数据文件格式，用于描述WARC文件中记录的Web内容，通常包含有关记录的URL、域名和其他元数据信息。
\item WET: WET文件通常包含从HTML页面中提取的文本内容，但不包括图像和其他资源。
\end{itemize}
CCNet使用的是WET数据构建文本数据集。图 \ref{CCNet}展示了其处理一个Common Crawl快照的流程，其包括：

\begin{itemize}
\item
\textbf{切分:}
如图 \ref{CCNet}所示，CCNet首先会把每个月的WET数据切分为5 GB的shard，一个月的Common Crawl大约会被切分为1到2000个shard。

\item
\textbf{去重:}
切分后，CCNet首先进行段落级别的去重：首先对文本进行处理，如段落所有字符小写、数字替换为placeholder、删除Unicode标点符号和重音符号等，然后每个段落做一次哈希，使用64-bits的SHA-1作为哈希值，保存为二进制文件中。CCNet在50个shard内进行去重，去重可以去除了网页导航栏、cookie警告、联系信息等。这可以从其他语言的网页中移除很大一部分的英文，对于后面识别语言减少了难度。

\item
\textbf{打分\&过滤:}
图 \ref{CCNet}中`LID'是`Language Identification'的缩写，表示一个基于fastText的语言识别模型。
对于每个网页，使用语言模型进行语言分类，得到分类分数。如果得分高于 0.5，则将该网页归类为特定语言，否则认为无法确定语言并丢弃该网页。
图 \ref{CCNet}中`LM'为`Language Model'缩写，使用KenLM库的 5-gram Kneser-Ney模型，用来进行语言质量打分。作者指出许多语言模型也能完成这项任务，但KenLM的效率更高。

\item
\textbf{分组:}
最后，根据语言和质量分数重新对文档进行分组，并写入JSON文件。shard会根据语言模型的困惑度（ppl）分数，将文档分为head，middle，tail三类，质量依次递减。

\end{itemize}

以上就是CCNet处理Common Crawl快照的具体流程。类似地，虽然其中一些环节的处理逻辑和细节可能不同，其他处理框架如SlimPajama、MNBVC、 RefineWeb\cite{nips/PenedoMHCACPAL23}等都基本遵循着切分、去重、打分、过滤、分组这一大体流程。
以现在的眼光看，CCNet作为一个比较经典的工作也存在着一些不足。首先，CCNet没有使用minhash，而是采用SHA-1，并且没有在整个数据集范围内进行去重。此外CCNet只关注一个文档整体的质量，没有粒度更小的质量判别机制，如句子质量判断等。
有很多后续的工作在处理网页数据时有着更严格和细致的处理。
例如，SlimPajama在CCNet基础上增加了整个数据集范围内的去重。
Falcon的RefineWeb增加了URL过滤机制，可以过滤掉一些欺诈、成人网站，同时使用了精确和模糊两种去重方法进行严格的去重。
此外，RefineWeb还增加了句子级别的过滤器，可以过滤掉一些无意义的句子。

\subsubsection{其他数据处理}

\begin{table}[t!]
\centering
\small
\caption{RedPajama-V2基于模型的质量注释}
\begin{tabular}{p{5cm}p{8.5cm}}
\hline

\textbf{指标名称} & \textbf{描述} \\
\hline
ccnet\_bucket & 困惑度评分：头部、中部、尾部 \\

ccnet\_language\_score & 语言识别模型的评分 \\

ccnet\_perplexity & 在维基百科上训练的语言模型的困惑度 \\

rps\_doc\_books\_importance & 文本相似于书籍数据的程度。 \\

rps\_doc\_openwebtext\_importance & 文本相似于OpenWebText数据的程度 \\

rps\_doc\_ml\_wikiref\_score & 基于维基百科引用文档的Fasttext分类器分数 \\

rps\_doc\_ml\_palm\_score & 基于维基百科，OpenWebText样本，RedPajama-V1书籍的Fasttext分类器分数 \\

rps\_doc\_ml\_wikipedia\_score & 基于维基百科文章的Fasttext分类器分数 \\
\hline

\end{tabular}
\label{model rules}
\end{table}

\begin{table}[t!]
\centering
\small
\caption{RedPajama-V2基于规则的质量注释}
\begin{tabular}{p{7.7cm}p{7.7cm}}
\hline

\textbf{指标名称} & \textbf{描述} \\
\hline
ccnet\_length & 字符数 \\
ccnet\_nlines & 行数 \\
ccnet\_original\_length & 文档内行重复删除前的字符数 \\
ccnet\_original\_nlines & 文档内行重复删除前的行数 \\
rps\_doc\_curly\_bracket & “{”和“}”出现的次数与原始文本中字符数的比例 \\
rps\_doc\_frac\_all\_caps\_words & 内容中仅由大写字母组成的词的占比 \\
rps\_doc\_frac\_lines\_end\_with\_ellipsis & 以省略号结尾的行的占比 \\
rps\_doc\_frac\_no\_alph\_words & 不含字母字符的词的占比 \\
rps\_doc\_lorem\_ipsum & “lorem ipsum”与字符数的比例 \\
rps\_doc\_mean\_word\_length & 词的平均长度 \\
rps\_doc\_stop\_word\_fraction & 停用词与词数的占比 \\
rps\_doc\_symbol\_to\_word\_ratio & 符号与词的比例 \\
rps\_doc\_frac\_unique\_words & 唯一词的占比 \\
rps\_doc\_unigram\_entropy & 文本的单元分布熵 \\
rps\_doc\_word\_count & 文本的词数 \\
rps\_lines\_ending\_with\_terminal\_punctution\_mark & 指示行是否以终端标点符号结尾 \\
rps\_lines\_javascript\_counts & 每行“javascript”出现的次数。 \\
rps\_lines\_num\_words & 每行的词数 \\
rps\_lines\_numerical\_chars\_fraction & 每行数字字符与字符总数的比例 \\
rps\_lines\_start\_with\_bulletpoint & 行是否以符号开头 \\
rps\_lines\_uppercase\_letter\_fraction & 每行大写字母与字符总数的比例 \\
rps\_doc\_num\_sentences & 句子数 \\
rps\_doc\_frac\_chars\_dupe\_5/6/7/8/9/10grams & 5/6/7/8/9/10grams重复比例 \\
rps\_doc\_frac\_chars\_top\_2/3/4gram & 频率最高2/3/4grams所占总字符比例 \\
rps\_doc\_ldnoobw\_words & 包含在LDNOOBW黑名单中的词序列的数量 \\
rps\_doc\_ut1\_blacklist & 文档领域的UT1黑名单分类ID \\
minhash\_signature\_0.7 & 文档的minhash指纹，用Jaccard相似度为0.7 /0.8 /0.9 /1.0的模糊去重 \\
\hline
\end{tabular}
\label{heuristics rules}
\end{table}

和网页数类似，预训练语料其他来源数据的处理大致分为去重、过滤和清洗三个步骤。

\begin{itemize}
\item
\textbf{去重:}
相比于网页数据，其他来源的数据通常总量更小，因此可以在整个数据集范围内进行去重，需要注意的是不同语言，不同来源数据需要设定不同的阈值，通常通过人工观察和验证来设定阈值。目前主流的去重方法为MinHash，其的典型工作流程如下:

\begin{itemize}
\item词袋生成（生成n-gram元组）后，将每个文档映射成一组哈希值（指纹生成），每个N-元组需要生成多个哈希值，再从哈希矩阵中取每一列的最小值（Min的来源），得到文档的MinHash值。

\vspace{10pt}

\item得到每个文档的MinHash值后，再做局部敏感哈希（LSH）。LSH指将指纹数组按行分成若干个条带 （band），每个条带的行数相同。若两个文档在某条带上MinHash值相同，这两个文档就会被聚到同一个桶中备选。逐条带（band）进行文档相似性的比较，并将相似的文档聚类可以大幅度减少后续比较的次数。

\vspace{10pt}

\item根据相似度决定保留或删除哪些文档。

\end{itemize}

\item
\textbf{过滤:}过滤指删除数据集中低质量的样本，其方法大体分为两种，一种是基于模型的方法还有一种是基于规则的方法。

\begin{itemize}
\item基于模型的方法通过训练质量打分/分类模型来过滤低质量数据。其使用高质量的预训练语料作为正样本，而将需要过滤的受污染文本作为负样本，来训练分类器进行有效过滤。
例如，The Pile使用训练了一个质量分类器来筛选低质量文本；除质量模型外，WanJuanText-1.0还专门训练了内容安全模型，用于检测并剔除与淫秽、暴力、赌博等主题相关的内容，确保语料库的健康和安全。

\vspace{10pt}

\item基于规则的方法通常是数据集构建者根据经验构建启发式的规则（heuristics rules），来判断一个数据样本是否应该被纳入数据集中。例如，RefineWeb通过 “总长度、符号与单词的比例以及一些衡量文本自然程度的指标” 来过滤异样本。
除了一些通用的启发式指标外，也有一些来源特定的启发式规则，例如，可以根据GitHub的Star数量过滤数据，移除没有Star或者Star数目过少的仓库；根据学术数据或书籍数据每行长度的分布筛选数据，移除那些可能是通过OCR （Optical Character Recognition，光学字符识别）得到的数据等。
\end{itemize}

参考CCNet，LLaMA，RefinedWeb， Gopher等工作，RedPajama-V2提供的40余种质量注释可以用来筛选和过滤数据。表 \ref{model rules}显示了其中基于模型的指标，表 \ref{heuristics rules}显示了其中基于规则的指标。

\begin{table}[t!]
\centering
\small
\caption{Flagdata\cite{flagdata}提供的部分文本清洗算子}
\begin{tabular}{p{5cm}p{8.5cm}}
\hline

\textbf{算子名称} & \textbf{描述} \\
\hline

CopyrightPruner & 清除代码样本中的版权声明 \\

EmailPruner & 清除文本样本中的电子邮件地址\\

IpPruner & 清除文本样本中的IPv4 和IPv6 地址\\

LinkPruner & 清除文本样本中的HTTP、HTTPS和 FTP链接 \\

NonChineseCharPruner & 清除文本样本中的非中文字符 \\

RepeatSentencePruner & 清除文本样本中的重复句子 \\

ReplacePruner & 将文本中符合特定正则表达式模式的所有内容替换为指定的替换字符串 \\

TablePruner & 清除文本样本中的表格文本\\

UnicodePruner & 规范化文本样本中的各种空白字符和Unicode标点符号 \\

FigureTableCaptionPruner & 清除文本样本中的图形和表格注释行 \\

ConsecutiveNewlinesPruner & 清除连续换行，允许最多两个连续换行 \\

CataloguePruner & 清除目录条目 \\

ControlCharPruner & 清除控制字符，保留换行符 \\

EndAtLastPunctuationPruner & 将文本截断到最后一次出现的标点符号 \\

SpecificPatternPruner & 根据提供的模式列表从文本中清除特定模式 \\

\hline

\end{tabular}
\label{baai opt}
\end{table}

\item
\textbf{清洗:}清洗指对文本样本进行修改，去除其中的噪声、冗余、有害、敏感信息等，清洗的过程可以在过滤之后也可以在过滤之前，视情况而定。过滤可以理解为样本级别的操作，即决定丢弃/保留一个样本，清洗则可以理解为样本内部的操作，即删除掉该样本中不需要的部分。数据清洗的目标包括：删除异常符号，广告信息/有害信息；百科代码数据的元信息，版权声明等；书籍和学术数据的封皮，版权页，目录等。社交媒体数据的按钮，导航栏，隐私信息等。
相比于文本过滤，文本清洗的大部分方法通常是基于规则的，通过统计分析方法和正则表达式实现。当然一些很难清洗的目标也可以通过训练小规模语言模型进行清洗。
智源研究院开源的FlagData数据处理框架提供了一些用于清洗文本的基础算子，表 \ref{baai opt}显示了其中一部分算子的名称和功能描述。

\end{itemize}

如本节所述，预训练处理大致分为去重，过滤，清洗三个部分。这三部分并没有固定的前后关系，有可能先去重再过滤，也可能先过滤再去重，当然也可能过滤、去重、再过滤。去重通常通过MinHash进行实现，通过迭代来减小需要比较的次数。过滤和清洗可以通过基于模型的方法和基于规则的方法完成。因为预训练数据的来源非常丰富且数据规模非常大，处理这些数据是一个相对复杂且工程量巨大的任务。目前很多公司和机构都在不同程度上开源了其数据处理流程，有一些还提供了可直接使用的工具仓库，如Together开源的RedPajama\footnote{https://github.com/togethercomputer/RedPajama-Data/tree/main}，
阿里开源的Data-Juicer\footnote{https://github.com/modelscope/data-juicer}，智源开源的FlagData\footnote{https://github.com/FlagOpen/FlagData}等，这很大程度地推动了数据建设研究的发展。








\subsection{国内外预训练数据集}
本节介绍目前主流的一些国内外预训练数据集。


\subsubsection{国内预训练数据集}

\begin{table*}[h]
\centering
\caption{国内预训练语料库}
\label{predata_cn}
\begin{tabular}{lcccc}
\hline语料库 & 大小 & 开源情况 & 开源协议\\
\hline
CLUECorpus2020 & 100GB & 开源 & MIT\\
WuDaoCorpora-Text & 5TB & 部分开源 & CC-BY-NC-ND-4.0\\
CCI 1.0/2.0/3.0 & 100GB/500GB/1TB & 开源 & CC-BY-NC-ND-4.0\\
WanJuanText-1.0 & 1.1 TB & 开源 & CC-BY-4.0\\
MNBVC & 32.3TB & 开源 & MIT\\
ChineseWebText & 1.42TB & 开源 &-\\
SkyPile & 600GB & 开源 & Skywork Community License \\
\hline
\end{tabular}

\end{table*}


表 \ref{predata_cn}显示了目前国内主要的预训练语料库，其详细介绍如下：

\begin{itemize}
\item
CLUECorpus2020\cite{bib20}是由CLUE组织发布的一个大规模中文语料库，它为自监督学习任务如语言模型的预训练或语言生成提供了丰富的数据资源。该语料库拥有100GB的原始文本数据，包含约35亿个中文字符，这些数据源自Common Crawl的网络爬取。它覆盖了新闻、娱乐、体育、健康、国际事务、电影和名人等多样化的主题，为中文NLP任务提供了广泛的应用场景。通过在不同规模的数据集上进行的语言理解实验，CLUECorpus2020证明了其训练出的模型在中文领域能够达到优秀的性能。

\item
WuDaoCorpora\cite{bib22}是由北京智源人工智能研究院构建的超大规模中文语料库，它为开发和训练大型预训练语言模型（PLM）提供了重要的支持。这个语料库包含了约3TB的训练数据，覆盖了教育、科技等50多个行业，为多领域预训练模型的训练提供了丰富的资源。WuDaoCorpora的数据经过了严格的清洗，并着重对隐私数据进行处理，避免隐私泄露风险。WuDaoCorpora的发布，不仅填补了世界范围内高质量超大规模中文语料库的空白，还为构建具有更强理解和生成能力的中文AI模型提供了坚实的基础。它的高质量数据和多模态特性，已经吸引了450多家产学研单位的研发团队下载使用，显著推动了中文自然语言处理技术的发展。

\item中文互联网语料库CCI\cite{baai_cci}是由智源研究院联合多家数据贡献单位构建的高质量、可信数据集，于2023年11月首次开源（CCI 1.0），旨在为国内大数据及人工智能行业提供一个安全、可靠的语料资源，共同推动大数据和人工智能领域的健康发展。在中国网络空间安全协会人工智能安全治理工作委员会、北京市网信办、北京市科委中关村管委会、海淀区等的推动下，智源研究院与多家企业经过四个月的努力，共收集约 8TB互联网数据，通过严格的清洗和筛选，形成了超过500GB的高质量“中文互联网语料库” CCI 2.0， 数据类型全面、质量较高、安全可信，包括网页、公众号、博客、百科、问答、试题等。

\item
WanJuan 1.0\cite{he2023wanjuancomprehensivemultimodaldataset}由上海AI实验室基于大模型数据联盟构建，经过细粒度清洗、去重和价值对齐，形成了具有多元融合、精细处理、价值对齐、易用高效四大特征的语料库。它包含文本、图文、视频三大数据集。其中文本部分总量超过1TB，覆盖科技、文学、媒体、教育、法律等领域，有效提升模型的知识含量、逻辑推理和泛化能力。通过语言甄别、正文抽取、格式标准化等精细化处理，以及数据质量评估，确保了数据的适配性和纯净度。此外，WanJuan 1.0采用统一格式，提供详尽的字段说明和工具指导，兼顾易用性和效率，适合快速应用于语言和多模态大模型训练。

\item
MNBVC\cite{bib23}语料集是一个全面而多样化的中文文本数据库，它不仅涵盖了主流文化内容，还包含了各种小众文化元素，甚至包括了火星文等特殊文本形式。该数据集汇集了新闻、作文、小说、书籍、杂志、学术论文、电影台词、网络帖子、维基百科条目、古诗、歌词、商品介绍、笑话、尴尬事件的叙述以及聊天记录等各类纯文本信息，所有数据均搜集自互联网。目前，MNBVC语料集的总数据量已达到33091GB，其目标是与chatGPT3.5的40TB数据量相匹配，现已完成了83\%的进度。

\item中文网络文本数据库（ChineseWebText）\cite{chen2023chinesewebtextlargescalehighqualitychinese}是由中国科学院发布的一个规模庞大、数据质量上乘的资源库，其总存储量达到1.42TB，只包含网络数据。其中，优质数据子集的存储量超过600GB，且数据质量评分高达90\%以上。该数据库为自然语言处理领域的研究者提供了一个极具价值的研究平台，特别适合用于训练和评估大型语言模型。该数据库不仅适用于深度学习模型的训练，而且在文本挖掘、情感分析、信息检索等多个领域均显示出广泛的应用潜力。对于企业级应用而言，中文网络文本数据库的数据资源可用于开发智能化搜索引擎、提高人工智能助手的理解能力，甚至构建高效的新闻摘要系统，以满足不同场景下的应用需求。

\item
SkyPile-150B\cite{skypile_150b}是昆仑万维专为大型语言模型预训练而设计的综合性大规模中文数据集，它基于大量可公开访问的中国互联网网页构建而成。为了保证数据的高质量，数据集经过了严格的过滤、广泛的重复数据删除以及彻底的敏感数据过滤。此外，研究人员运用了fastText和BERT等先进工具，进一步筛选出低质量数据，确保了数据集的纯净度和可靠性。公开部分的SkyPile-150B数据集包含了约2.33亿个网页，平均每个网页包含超过1000个汉字，总计约1500亿个Token，其纯文本数据量达到了620GB。
\end{itemize}

\subsubsection{国外预训练数据集}
表 \ref{predata_f} 展示了目前国内主要的预训练语料库，详细介绍如下

\begin{table*}[h]
\centering
\caption{国外预训练语料库}
\label{predata_f}
\begin{tabular}{lcccc}
\hline
\textbf{语料库} & \textbf{大小} & \textbf{开源状态} & \textbf{开源协议} \\
\hline
RedPajama-1T & 5TB & 开源 & Apache-2.0 \\
SlimPajama & 895 GB & 开源 & Apache 2.0 \\
RedPajama-V2 & 180TB & 开源 & Apache-2.0 \\
The Pile & 825.18 GB & 开源 & MIT \\
Refinedweb & 5T & 部分开源 & ODC-BY-1.0 \\
FineWeb & 44T & 开源 & ODC-BY-1.0 \\
TigerBot pretrain en\textbackslash zh & 51 GB\textbackslash 55 GB & 部分开源 & Apache-2.0 \\
Dolma & 11.24 TB & 开源 & MR Agreement \\
C4 & 806.87 GB & 开源 & Apache-2.0 \\
MassiveText & 10.5 TB & 未开源 &-\\
ROOTS & 1.61 TB & 开源 & BLOOM Open-RAIL-M \\
OSCAR & 8.41 TB & 开源 & CC0 \\
OpenWebMath & 26 GB & 开源 & ODC-BY-1.0 \\
\hline
\end{tabular}
\end{table*}

\begin{itemize}
\item RedPajama-Data-1T（RedPajama-1T）\cite{redpajama_data_1t}是Together主导，与Ontocord.ai、ETH DS3Lab等多个公司与组织合作发布的一个预训练数据集。是RedPajama项目发布的第一组训练数据。RedPajama-1T涵盖了六种不同的数据类型，来源包含CommonCrawl、C4、GitHub、arXiv、书籍、维基百科、StackExchange等7个来源，并经过了仔细的数据预处理和过滤，其数据配比大致模仿Meta AI在LLaMA论文中报告的token数量， 共1.2T tokens， 其中来自CommonCrawl的网页数据占比达到87.0\%。RedPajama不仅开源了其预训练数据，而且开源了其数据处理方法并开提供所有数据预处理和质量过滤器，再业界和研究领域内收到了巨大欢迎，发布半年后已经有超过190 000的下载量。其不仅提供了宝贵的研究资源也推动了大语言模型时代开源社区的构建。

\item SlimPajama\cite{shen2024slimpajamadcunderstandingdatacombinations}数据集是由Cerebras基于RedPajama-1T数据集通过深度去重得到的英文数据集。因为RedPajama-1T处理数据时主要关注一条数据本身的质量，并没有在整个数据集范围内进行广泛的去重。针对这一点，Cerebras开发了一个包含精确和模糊算法的全量去重框架。Cerebras先删除短小且质量较低的文档，然后利用精确（13-Grams）和模糊（MinHashLSH），通过分布式并行和内存池优化方法实现了对 1.2T Tokens的全量去重，滤掉了 49.6\% 的重复Tokens，得到 0.63 T tokens。其汇报的实验结果表明，使用SlimPajama训练的模型性能优于使用RedPajama-1T数据集训练的模型。

\item RedPajama-V2\cite{bib16}数据集是Together等在RedPajama-Data-1T后发布的第二版的预训练数据集。该数据集专注于Common Crawl的网页数据，处理了大量的84个Common Crawl快照，包含1000亿份文本文档，原始Token数目高达 100T。处理后的数据涵盖5种语言（英语、法语、西班牙语、德语和意大利语），Token数目为30T， 是之前版本的接近30倍。此外，除了使用CCNet pipeline对网页数据进行处理外，RedPajama-V2还提供了40多个预先计算的质量注释，包含“不包含字母字符的单词比例”等启发式注释和使用机器学习方法获得的质量分数等，这些注释有助于使用者进一步筛选和加权数据。RedPajama-V2 为研究者和开发者提供了一个低成本获取大量网页数据的途径，并且其注释也有助于对不同的网页数据对模型训练的影响进行分析。

\item The Pile\cite{gao2020pile800gbdatasetdiverse}是由Eleuther AI社区构建的大规模英文数据集，其规模达到 825GB，包含约 1.4 亿份文档，平均每个文档大小为 5.9 KB。The Pile有着广泛的数据覆盖范围，包含五个主要类别：学术材料、网页、书籍、对话和代码。The Pile由Pile-CC，PubMed Central，Books3等22个不同的高质量子集组成，各个子集在构建过程中采用了“分治”思想，根据数据特征定制清洗规则。The Pile作为一个多来源预训练数据集，其特点为关注书籍，学术资料等数据，网页数据占比较小。

\item Refinedweb\cite{nips/PenedoMHCACPAL23}是阿布扎比技术创新研究所（Technology Innovation Institute， TII）开发的数据集，该数据集只包含网页数据，专为大规模模型预训练而设计。
RefinedWeb从Common Crawl（CC）的网页数据中经过精心的清洗和筛选而形成，经过清洗后的文本的淘汰率接近90\%。
Refinedweb仅包含网页数据，其构建者声称，只依靠经过严格筛选的网页数据进行训练，语言模型的表现也能与那些使用多种专有数据混合的方案相比肩。
使用RefinedWeb数据集已经成功训练了多个模型，包括Falcon-7B、40B和 180B，Refinedweb构建者声称，RefinedWeb的数据质量超过了其他数据集，如RedPajama-1T和 SlimPajama。
Refinedweb总共包含了 5T Tokens，其中 600B Tokens已向公众开源。

\item FineWeb\cite{penedo2024finewebdatasetsdecantingweb}是 Hugging Face提供的一个大规模英语网页数据集，包含超过 15 万亿个经过清洗和去重的token，源自CommonCrawl。该数据集的初衷是制作一个完全开源的RefinedWeb的复制，但是构建者在构建过程中加入了很多精心设计的过滤步骤。作者声称使用FineWeb数据集上训练的模型比使用其他网页数据集（如C4，Dolma-v1.6，The Pile，SlimPajama，RedPajama2）训练的模型效果要好。FineWeb旨在提供一个完全开源的高质量网页预训练数据集，并优化大型语言模型（LLM）的训练数据处理流程，其使用datatrove库进行处理。

\item TigerBot pretrain\cite{chen2023tigerbotopenmultilingualmultitask}是Tiger基于GPT3 的pretrain的数据分布，采集中文书籍，互联网，和百科类数据，从 20TB数据清洗过滤到 2TB，同时保持了语言和类目的比例，
该组织基于该数据集训练了TigerBot family系列模型并进行了开源。TigerBot pretrain已经开源了其中的100GB数据。

\item Dolma\cite{bib408}
是AI2最新发布的英文预训练数据集，用于训练开源大模型OLMo等。Dolma有 3T tokens，涵盖了网页内容、学术出版物、编程代码、书籍和百科全书等多种类型的数据。
Dolma是AI2发布的最大规模数据集，信息透明度也非常高，AI2官方发布了非常详细的处理流程。此外，Dolma采用了非常友好的开源协议，支持免费商用，以促进人工智能技术的发展和应用。

\item MassiveText\cite{bib52}是DeepMind开发的一个大型的英文数据集，包含大约 23.5 亿份文档，总数据量接近 10.5TB。这个数据集涵盖了网页、书籍、新闻文章和代码等多种内容，但并没有被开源。尽管如此，MassiveText数据集的构建过程已被公开，为学术界和开发者提供了制作类似数据集的指导。这个数据集用于训练具有 280B参数的Gopher模型。

\item C4（Colossal Clean Crawled Corpus）\cite{raffel2020exploring}是Google基于Common Crawl创建的大规模英文语料库，含有大约 3.65 亿份文档（156B Tokens）。这个数据集用来训练了T5 和Switch Transformer模型。在C4 数据集的构建过程中，研究者们采用了多种数据清洗启发式规则，比如只保留以终止标点结束的文本行、丢弃少于5个句子的网页，同时每个句子需至少包含 3 个单词等。这些清洗规则为后续相关研究提供了宝贵的经验。C4被用来训练Google的 T5和LaMDA，以及Meta的 LLaMA大型语言模型。

\item ROOTS\cite{bib42}是BigScience团队开发的一个庞大的多语言数据集，总大小为 1.6TB。这个数据集支持了BLOOM模型的训练，该模型具有超过 1700 亿个参数。ROOTS涵盖的语种多达 59 种，包括 46 种自然语言和 13 种编程语言。
在其数据来源上，ROOTS数据集有62\% 的数据由社区进行搜集和筛选；另外 38\% 的数据则来自OSCAR数据集。
为确保高质量，ROOTS的数据清洗采用了严格的清洗标准，并使用Simhash算法进行文本去重。此外，ROOTS数据集还进行了个人信息的剔除，以符合数据保护和隐私的规范。

\item OSCAR（Open Super-large Crawled Aggregated coRpus）\cite{SuarezRS20}是一个开放源码项目，旨在从Common Crawl的 WET文件中提取数据，为多语言的词嵌入和大规模语言模型预训练提供基于网页的数据集。该项目已成功发布了一个包含 151 种语言的多语言数据集OSCAR，总大小为 6.3TB，最新版为OSCAR 23.01。研究人员通过使用基于OSCAR数据库的ELMo上下文词嵌入模型进行训练，验证了其清洗算法在减少Common Crawl数据噪声方面的效果，证实数据质量可与Wikipedia相媲美。此外，OSCAR项目还开发了一个用于分类和过滤Common Crawl数据的高效并行处理框架，名为Goclassy和Ungoliant。该框架利用FastText的线性分类器及其预训练模型来进行语言识别。

\item OpenWebMath\cite{paster2023openwebmathopendatasethighquality}是一个数学相关的预训练语料，语料来源为Common Crawl的数学网页。
OpenWebMath构建的关键是从Common Crawl庞大的网页中过滤出数学相关的网页，其将数学网页定义为包含核心数学内容的文档，例如定理、定义、证明、数学公式、问题和答案、形式数学等。OpenWebMath使用一种pipeline的方式过滤网页：1. 检查常见的数学字符串（如math，mathjax）、2. 搜索LATEX符号、 使用MathScore分类器打分。最终在超过200B的HTML文档中提取了14.7B Tokens作为训练语料。

\end{itemize}


\subsection{小结}
大规模预训练数据集的构建是训练大语言模型的基础，近年来受到广泛关注。许多公司和组织投入大量资源参与到语料数据的构建和优化工作中，以提升训练出的语言模型的性能。

在数据来源层面，网页数据方面，英文网页的数量和质量明显高于中文网页，英文网页数据在全球仍处于主流位置，这为英文大语言模型的训练提供了丰富的资源。此外，自媒体和智能手机的兴起很大程度的改变了人们的上网习惯，网页端的内容关注度在逐步减少，每年都会有大批的网页消失，如何及时保存现有的中文网页资源是一件非常值得关注的事情；
学术数据方面，英文数据也占据主流的位置，能获取的开源语料大部分都是英文撰写，开源的中文学术语料库数量有限；
书籍数据方面，中文数据的劣势相对较小，据不完全统计，可获得的中英文书籍数据的体量大概在1比5左右；
社交媒体数据方面，中文和英文数据都相对闭源。大多数社交媒体平台对数据的获取设置了限制，运营商之外的组织和公司很难获取这些数据来构建训练数据集。这在一定程度上限制了大语言模型在社交媒体语料上的训练和优化。考虑到近年来自媒体的兴起，很多新产生的高质量语料都会出现在微信公众号，小红书等社交媒体平台而不是公开的网页。社交媒体数据的互联互通可以有效增加预训练语料库的多样性和规模，然而，在大数据时代，数据是重要资源，也是核心竞争力，这一方向的推进前景并不乐观。

在原始数据处理层面，业内普遍遵循相似的数据处理框架，但大公司的数据处理流程公开的信息较为粗略，缺乏细节描述，如去重的粒度和过滤清洗的具体规则等。相比之下，许多研究机构和非盈利组织开源了详细的数据处理流程和工具，并基于这些数据训练了高质量的语言模型。这为研究者们提供了宝贵的经验和实践指导，为开源社区做出了巨大贡献。

在开源数据集层面，中文数据集在规模和质量上与英文数据集存在较大差距。一个重要原因是中文网页数量少且高质量中文数据的获取成本较高。例如，Common Crawl中的中文网页占比只有约1.4\%，远低于英文网页的55\%。尽管国内的大规模语言模型（如Baichuan、Qwen、ChatGLM等）已经相继开源，但其预训练数据几乎全部闭源。这导致在中文领域，高质量大规模开源预训练数据仍然处于一个相对缺失的状态。
虽然很多训练经验表明，即使使用以英文数据为主的预训练数据，语言模型仍然可以表现出较强的多语言能力。例如，ChatGPT的训练数据中英文数据占了百分之九十以上，但使用者仍可以使用中文与其流畅对话，甚至能够识别中文俗语，并根据风格生成不同的中文文本。然而，ChatGPT的中文能力仍明显低于其英文能力。例如，同一个问题使用英文提问获取正确答案的概率比使用中文提问更高，英文回答通常更长且描述更为详细。同时，中文领域特有的知识、中文文本风格、文言文的理解和书写并不能通过多语言之间的泛化实现。因此，构建高质量的中文预训练语料非常必要。
构建一个包含多样化、高质量的中文预训练数据集，将有助于提升中文大语言模型的整体表现，使其在各种应用中更具竞争力。
这需要各方的共同努力。例如，可以通过跨领域的数据共享和合作提升中文预训练语料库的数据量和多样性。

此外，随着语言模型使用的训练数据量的膨胀，数据隐私问题得到了广泛关注。如何在保障用户隐私的同时提供高质量、可用的数据是一个值得关注的问题。预训练语料的构建需要在数据开放性和隐私保护之间找到平衡点。
数据开放性与隐私保护之间的平衡涉及多个层面。首先，数据提供者和数据使用者需要建立信任关系，确保数据在使用过程中不会被滥用。其次，需要制定和遵循严格的数据隐私保护政策和法规，确保数据处理过程透明且合规。最后，技术手段也需要不断改进，如数据匿名化技术、差分隐私技术等。

最后，大规模预训练数据的构建搜集了网络上的大部分语料资源并进行了处理，这也导致研究者们对语料枯竭的担忧。如果没有新的语料来源，预训练数据集的质与量难以得到本质提升，那么大语言模型的性能是否会达到一个较难突破的上限呢？为了应对这一挑战，研究者们试图探索现存数据以外数据来源（合成数据），
以确保预训练数据集的持续改进和优化，这是一个比较新且值得关注的研究方向。



\section{微调数据}


微调过程是训练语言模型不可或缺的一环，其目的在于将模型从面向语言知识的通用训练转向面向具体任务的针对训练。在预训练阶段，模型通过预测被掩码的Token任务学习了大量的语言知识，使其具备理解和生成文本的能力。然而，这些知识是通用的，无法直接应用于具体任务中。微调通过让语言模型在针对特定任务的语料库上进行训练，使模型能够学习到具体任务的需求和规律，从而将所学的语言知识和生成能力用在解决特定任务上。
与预训练数据不同，微调数据集通常是经过人工标记的语料数据，这些数据直接与具体的应用任务相关。例如，对于情感分析任务，微调数据包括文本和该文本情感类别的标签；对于阅读理解任务，微调数据包括参考文本，问题和该问题的答案。通过在这些针对专门任务的数据上训练，模型能够学习到解决特定任务的特征和模式，从而将预训练阶段学到的通用知识应用到特定任务中。

传统的自然语言处理任务包括：命名实体识别，机器翻译，文本分类，自动问答等。这些任务需要语言模型有着多方面的能力，如理解能力，推理能力，上下文关联能力等，并分别有所侧重，如文本分类主要考验模型的语言理解能力，自动问答考验模型的理解，知识储备和生成能力等。
然而，随着ChatGPT的发布，人们发现，如果一个语言模型在预训练阶段学习了足够多的语言知识，获得了足够强的语言能力，便能以对话的模式解决几乎所有的自然语言处理任务，使得大规模语言模型获得前所未遇的通用性。在这之后训练通用性强的大规模语言模型成为了主流，模型的参数量，使用的预训练数据集大小在规模上产生了巨大的膨胀，而微调数据集的变化则主要体现在其形式和难度上：
\begin{itemize}
\item传统微调数据集可能有着多种形式：如文本-文本（机器翻译），文本-类别（文本分类），文本-span范围（抽取，识别）等。但是在大规模语言模型时代，微调数据集统一为文本-文本的形式，且通常以问题/指令-回答这种对话模式存在。这种对话的形式可以解决的问题非常广泛，并且有着更好的交互性。

\item大规模语言模型使用的微调数据集在任务难度上相对于传统微调数据集更为困难，需要模型有着更强的知识储备，推理能力，文本理解能力和生成能力。例如，传统的问答数据集经常需要给定模型一段回答问题所需的参考文本，而现在的微调数据集更倾向于直接向模型进行提问，让模型根据自己掌握的知识进行回答。此外，现在的微调数据集还会关注数学推理，代码生成等较为复杂的任务。
\end{itemize}

\begin{table}[htb]
\centering
\caption{指令跟随数据集Alpaca中的两个示例}
\begin{tabular}{p{5cm}p{9cm}}
\hline
\textbf{指令} & \textbf{回答} \\ \hline
 我们如何减少空气污染？ & 
减少空气污染的方法有很多，例如转向可再生能源、鼓励使用公共交通、禁止燃烧化石燃料、实施减少工业排放的政策以及实施车辆排放标准。此外，个人也可以通过减少汽车使用、避免燃烧木材等材料以及改用节能电器来减少空气污染。 \\
& \\

 给出三条保持健康的秘诀。 & 
1. 均衡饮食，确保摄入大量水果和蔬菜。\\
& 2. 经常锻炼，保持身体活力和强健。\\
& 3. 保证充足睡眠，保持规律的睡眠时间。\\ 
 \hline
\end{tabular}
\label{instruction-following}
\end{table}

\begin{table}[htb]
\centering
\caption{模型微调前后生成对比}
\begin{tabular}{p{5.5cm}p{4cm}p{4cm}}
\hline
\textbf{指令} & \textbf{微调前回答} & \textbf{微调后回答} \\ \hline根据Yelp评论预测作者会给餐厅打多少颗星：“老实说，我觉得这家餐厅像一个打卡景点。食物很一般，但是服务还可以。面包很好吃。我点了薯条，很软，味道也不好。男朋友点了一份香蒜鸡肉意面，味道也很一般。” &
食物还可以，但不是很好。服务还可以，但不是很好。氛围还可以，但不是很好。价格还可以，但不是很好。& 根据作者的评论，他会给这家餐厅打两颗星。 \\

\hline
\end{tabular}
\label{instruction-following2}
\end{table}

微调数据集的一种常见任务是一种叫做指令跟随（instruction-following）的任务，这个任务旨在训练模型跟随人类指令的能力，即让模型把学到的知识和语言能力应用于解决人类提出的问题或要求。
表 \ref{instruction-following}展示了指令跟随数据集Alpaca的两个示例\footnote{显示翻译为中文后的内容}。表 \ref{instruction-following2}展示了一个base模型（只预训练）在Alpaca数据集\cite{taori2023stanford}上微调前和微调后针对一个问题的回答。可以看到，微调前的模型只有语言生成的能力，但是并不能帮助使用者解决提出的问题，微调后的模型可以较好地回答用户提出的问题。需要注意的是Alpaca数据集只有52k，这说明模型并不是在如此小规模的数据上学习到解决这些问题的知识，而是模型学习到了应该以回答问题的方式进行语言生成。所以说指令跟随是一个将模型学习到的语言知识应用在人类需要解决的任务上的一个过程。
可以看到微调数据集的构建是构建大规模语言模型的一个关键步骤，直接影响到模型解决任务的能力，决定了模型与人类交互的最终体验。
本章介绍用于训练大规模语言模型的微调数据集，其主要任务包括指令跟随、日常对话、数学推理、代码生成、医疗问答等。构建微调数据集的主要问题分为两个方面，问题的收集和答案获取。其构建方式大致可以分为以下几种：人工撰写、真实用户对话、基于已有数据集的扩充、利用语言模型直接生成以及基于传统数据集的改写。下面介绍这几种基本的数据构建方式。

\subsection{人工撰写}

人工撰写指令和答案是获取高质量微调数据集最直接的方法。
人工构造的数据集可以通过充分的培训和审查，保证数据的准确性和相关性，提供高质量的训练数据。
例如，在OpenAI就选择了外包标注人员编写的指令和回答作为训练ChatGPT 3.5的微调数据。
OpenAI通过一系列的筛选，找到了40个对不同人口群体的偏好敏感并且善于识别可能有害的输出的全职标注人员（约89\%持有本科及以上学历）。
标注人员的工作是根据要求自己编写指令，指令分为通用问答、头脑风暴、分类任务、生成任务、聊天等多种类型。数据构建过程中，标注人员被要求完成以下三个工作内容：
\begin{itemize}
\item要求标注人员给出符合指令类型的简单指令，需要确保任务的多样性。
\item要求标注人员写出同一个指示的多种不同的表达。
\item要求标注人员根据测试用例编写指令\footnote{OpenAI训练了一个早期版本的InstructGPT，开放给了一部分用户并收集了使用数据。}。
\end{itemize}

在整个过程中，OpenAI的研发人员跟这些标注人员紧密合作，给他们进行了培训，并对他们的工作质量进行评估，以确保他们能够代表大众的偏好并提供高质量的回答。
根据OpenAI的技术报告\cite{openai2022introducing}，其最终构建了一个13k个由标注人员编写的指令及其对应回答的SFT数据集用来微调预训练模型。

人工构造的微调数据集有着较高的语言质量，生成答案准确且细节可控，能够满足严格的要求（例如答案的长度范围），并可以确保内容的安全性与无害性，且不存在幻觉问题。
能提供高度精确且关联性强的训练样本，极大地促进了模型的准确性和可靠性。
然而，人工构建数据集成本较高，为保证数据质量，需要对外包人员进行培训和和持续的跟踪。此外，人工构造微调数据集的重要不足在于通过人工构建的指令很难全面覆盖真实使用场景的多样性和复杂性，这种局限性可能会导致模型在应对未见过的或非常规问题时表现不佳。
为了缓解这一问题，OpenAI采取了设计多样化的任务类型和引入InstructGPT\cite{ouyang2022training}用户的真实指令作为参考。
尽管这种方法在一定程度上增强了数据集的全面性，但要完全覆盖广泛且多变的真实场景仍是一大挑战。

表\ref{generated_data1} 列出了一些人工撰写的数据集和其基本信息，这些数据集的详细介绍如下：

\begin{table*}[t!]
\caption{人工撰写的数据集}
\centering
\label{generated_data1}
\begin{tabular}{lcc}
\hline
\textbf{数据集名称} & \textbf{样本数量} & \textbf{许可证} \\
\hline
databricks-dolly-15K & 15K & CC-BY-SA-3.0 \\
OASST1 & 161K & Apache-2.0 \\
Aya Dataset & 204K & Apache-2.0 \\
LIMA & 1K & CC BY-NC-SA \\
MATH & 12K & MIT \\
GSM8K & 8K & MIT \\
\hline
\end{tabular}
\end{table*}

\begin{itemize}
\item databricks-dolly-15K\cite{bib60}是Databricks公司于2023年4月发布的开源数据集，包含15 011个人工生成的高质量指令-回答对。该数据集涵盖了开放式问答、分类、情感分析、创意写作、代码生成等多个类别，旨在训练和评估大型语言模型指令跟随能力。其由Databricks的5000名员工人工生成，确保了数据的多样性和质量。这个数据集适用于商业和企业应用场景，可用于微调大型语言模型以执行特定任务，如客户服务自动化、内容生成和数据分析等。

\item OASST1\cite{OpenAssistant_oasst1}是OpenAssistant项目在2023年4月发布的大规模多语言对话式AI数据集，包含161 443个人类编写的提示、回复及相应评分。支持35种语言，涵盖了从日常对话到专业讨论的广泛主题。数据集的创建采用了众包方法，确保了内容的多样性和真实性。OASST1的主要目标是训练能够理解和生成多语言、多轮对话的AI助手，特别适合开发跨语言、跨文化的对话系统。它在提升模型的语言理解、上下文把握和回应生成能力方面具有重要作用，可应用于多语言客服、跨文化交流辅助工具等领域。

\item Aya Dataset\cite{bib409}是由Cohere For AI Community等机构于2024年发布的一个开源、多语言的指令风格数据集。该数据集包含204 114个指令-回答对，覆盖65种语言，旨在填补现有指令跟随数据集中的语言差距。Aya数据集通过Aya Annotation Platform人工收集和标注，涉及2997名来自119个国家的合作者。该项目不仅发布了Aya数据集，还包括Aya Collection（包含44个指令风格数据集）和Aya Evaluation Suite（涵盖13种语言的人工标注样本），以及将高质量数据集翻译成101种语言的Aya Collection。总计，这些资源组成了目前最大的多语言指令微调数据集，支持对低资源语言的研究和数据创建。

\item LIMA\cite{bib90}是Meta AI等机构发布的一个数据集，仅包含1000个实例，用于大语言模型的指令微调研究。它的特点是仅使用1000个精心策划的提示和回应对65B参数的LLaMa模型进行微调，不使用强化学习或人类偏好建模。研究发现，即使只用这么少量的数据，模型也能展示出强大的性能，能处理复杂查询并具有良好的泛化能力。这项研究提出了“表面对齐假说”，认为模型的核心能力主要在预训练阶段获得，而对齐过程则教会模型如何以合适的方式与用户互动。

\item MATH\cite{bib334}包含12 500个具有挑战性的竞赛数学问题的数据集。每个问题都有一个完整的循序渐进的解决方案，可以用来教授模型生成问题的推导过程和答案。
MATH的问题来源于AMC 10、AMC 12、AIME等数学竞赛。这些竞赛被用于评估优秀的年轻数学人才解决数学问题的能力。

\item GSM8K\cite{cobbe2021training}由OpenAI发布，包含8.5K小学数学问题，分为7.5K的训练集和1K的测试集。GSM8K提供问题的自然语言解决方案，每个问题的解题步骤在2-8步，每步为普通的加减乘除。GSM8K数据集的设计遵循了以下原则：高质量、高多样性、适中难度以及自然语言解决方案。这些问题由人类问题编写者精心创建，并通过广泛的质量控制措施来确保其质量。

\end{itemize}

\subsection{真实用户数据}

真实用户数据是基于用户与用户或用户与AI系统之间的实际交互收集而来。
由于它基于真实的用户需求和反馈，因此可以反映用户在现实场景中表达需求的语言使用方式，包括查询方式、语言的自然流畅性以及交互过程中可能出现的各种语言模式等。其优点在于：
\begin{itemize}
\item自然性：对话来自真实用户的自发查询，未经过编辑或简化，包含自然语言的特有复杂性和多样性。
\item广泛性：数据来源于自世界范围各个领域阶层的用户，反映了不同群体的需求和偏好。
\end{itemize}

真实的用户数据对于改进语言模型非常宝贵，可以帮助模型更好地理解和响应人类用户。
然而，与人工构建的数据相比，真实的用户对话可能包含噪声，如无效对话和有害信息。因此基于真实用户对话构建微调数据集需要对收集的数据进行额外处理，如过滤掉无效对话和有害/无关内容、修正错误信息，确保数据集的准确性和可用性。
此外，因为用户的需求分布可能是不均匀的，可能还需要对数据进行分析，去掉类似的样本，尽可能的保证数据集的多样性。
值得注意的是，由于使用来自于用户于模型之间的真实对话，在处理和使用这些对话数据时，需要确保用户的隐私得到保护。通常需要对数据进行脱敏处理，去除个人隐私信息。实际上，利用用户与AI系统之间的交互数据进行模型训练可能触及用户权益的敏感问题。例如，用户是否知晓并同意他们的使用数据被记录并用于模型训练或数据集构建。在任何情况下，构建数据集和训练模型都必须严格遵守当地的法律法规并防止数据滥用的情况发生。

一些公司如OpenAI在大规模语言模型上具有先发优势，通过向用户提供对话模型，收集积累了大量来自全球用户的真实人机交互数据，其可以利用这些数据不断地优化其微调数据集和改进模型。例如，ChatGPT、文心一言等模型在提供回答后会提供用户评价回答好坏的选项。通过这种用户反馈，可以识别模型的错误回答或无法正确解决的问题，然后针对性地构建和迭代微调数据集。这种持续的用户反馈给头部模型持有者带来了宝贵的数据积累和较为明显的竞争优势。

表\ref{generated_data2} 列出了一些基于真实用户数据构建的数据集和其基本信息，这些数据集的详细介绍如下：

\begin{table*}[htb]
\caption{基于真实用户数据构建的数据集}
\centering
\label{generated_data2}
\begin{tabular}{lcc}
\hline
\textbf{数据集名称} & \textbf{样本数量} & \textbf{许可证} \\
\hline
InstructionWild\_v2 & 110K & MIT \\
LCCC & 12M & MIT \\
Chatbot\_arena\_conversations & 33K & CC-BY-NC-4.0 \\
Zhihu-KOL & 1M & MIT \\
Chinese-Medical-DIALOGUE-Data & 792K & MIT \\
OpenChat & 70K & MIT \\
ShareGPT-Chinese-English-90k & 90K & Apache-2.0 \\
HuatuoGPT-SFT-data-v1 & 226K & Apache-2.0 \\
Huatuo-26M & 26.5M & Apache-2.0 \\
LMSYS-Chat-1M & 1M & LMSYS-Chat-1M license \\
WildChat & 1.04M & AI2 ImpACT license \\
MMIQC & 2.3M & Apache-2.0 \\

\hline
\end{tabular}
\end{table*}

\begin{itemize}
\item InstructionWild\_v2\cite{bib62}是新加坡国立大学发布的一个大规模指令数据集。该数据集包含11万条多语言（英文和中文）指令，来源于网络上收集的ChatGPT使用分享。与此前的小规模人工编写指令集或基于少量种子指令生成的数据集不同，InstructionWild\_v2提供了更大规模、更多样化的真实指令样本。研究者还基于这些指令生成了52K条指令-回复对。该项目旨在为指令调优研究提供更丰富的开源数据资源，以推动大型语言模型的发展。

\item LCCC\cite{WangKZHJZH20}是一个由清华大学等机构发布的大规模中文对话数据集。它包含基础版（680万对话）和大型版（1200万对话），数据来源于社交媒体上的用户互动。LCCC通过严格的数据清洗流程保证了高质量，主要用于训练短文本对话生成模型。该数据集及其预训练模型均已在GitHub上开源，采用MIT许可证。LCCC的发布为中文短文本对话建模研究提供了宝贵的大规模高质量语料资源，有望推动相关领域的发展。

\item Chatbot\_arena\_conversations\cite{chatbotarena}数据集由加州大学伯克利分校等机构发布，包含33 000个清理后的对话实例，并配有人类偏好对比标签。该数据集涵盖从2023年4月至6月在Chatbot Arena上收集的13 000个独立IP地址生成的对话。每个样本包括问题ID、两个模型名称、完整对话文本（OpenAI API JSON格式）、用户投票、匿名用户ID、检测到的语言标签、OpenAI审核API标签、额外的有害标签和时间戳。数据集的发布遵循CC-BY-4.0和CC-BY-NC-4.0许可，并通过数据收集网站的“使用条款”获得用户同意。尽管数据集发布方尽力移除所有包含个人身份信息（PII）的对话，但保留了不安全对话，以供研究人员研究大语言模型（LLM）在真实世界应用中的安全问题及OpenAI审核过程。此外，数据集中包含由T5和RoBERTa模型微调的有害标签器生成的附加有害标签。

\item Zhihu-KOL\cite{wangrui6_zhihu_kol}是一个从中国知名问答平台知乎爬取的大规模中文数据集，包含1 006 218个高质量问答内容，特别是来自“知乎专栏”的文章。数据集涵盖了广泛的主题，包括科技、文化、社会、经济等多个领域。它的独特之处在于内容质量高，多为知乎平台上的“大V”（关键意见领袖）撰写的专业回答和文章。这个数据集对训练中文语言模型和问答系统极具价值，可用于改善模型在专业领域的表现，提升长文本理解和生成能力。它特别适合用于开发垂直领域的问答系统、知识图谱构建和自动文章摘要等任务。

\item Chinese-Medical-DIALOGUE-Data\cite{BillGPT_chinese_medical_dialogue}数据集由开发者Toyhom创建并维护，旨在为医疗领域的自然语言处理研究提供丰富的资源。该数据集包含了大量真实的医疗对话记录，涵盖疾病诊断、用药指导等多个主题，并附带结构化的标签信息，如病症、药物、症状等。数据集具有较高的真实性和多样性，特别适合开发高质量的医疗对话机器人。项目遵循MIT许可证，允许自由使用、修改和分发，并保持活跃更新。

\item OpenChat\cite{WangCZLSL24}数据集是由清华大学等机构发布的多轮对话数据集，包含约7万个来自ShareGPT的英语对话实例。这是一个混合质量的数据集，其中约6000条是由GPT-4生成的高质量专家对话，其余则是GPT-3.5生成的次优质量对话。其独特的混合质量特性可能有助于训练出更加鲁棒和多样化的模型。

\item ShareGPT-Chinese-English-90k\cite{ShareGPT-Chinese-English-90k}是一个高质量的中英文双语人机对话数据集，包含约9万条真实用户场景下的问答对。这个数据集的主要特点是提供完全对应的中英文平行语料，适合训练双语对话模型；所有问题来源于真实用户提问，而非人工构造或API生成，因此更符合实际用户的指令分布和表达方式；数据通过网友自发分享收集，经过了一层自然的人类筛选，保证了对话质量。这些特性使得该数据集在训练高质量、鲁棒性强的对话模型方面具有显著优势，特别是相比于那些通过API接口生成的模拟问答数据。

\item HuatuoGPT-SFT-data-v1\cite{FreedomIntelligence_huatuogpt_sft_data_v1}数据集由香港中文大学等机构发布，包含226 042个实例，遵循Apache-2.0许可。该数据集来源于医生和患者之间的真实对话以及由ChatGPT生成的对话，涵盖多种指令类别和医疗领域。其设计目标是通过结合真实世界数据和人工生成数据，使模型具备医生般的专业性和患者友好的特点。数据集分为四种类型：从ChatGPT中提取的指令、医生的真实指令、从ChatGPT中提取的对话和医生的真实对话，每种类型都包含指令数据和对话数据，以增强模型的指令跟随能力和互动诊断能力。

\item Huatuo-26M\cite{abs-2305-01526}是目前最大的中文医疗问答数据集，由香港中文大学等机构发布，包含26 504 088个实例，遵循Apache-2.0许可。数据集从多个来源收集和整合，包括在线医学百科全书、在线医学知识库和在线医疗咨询记录，涵盖疾病、症状、治疗方法和药物信息等多个方面。Huatuo-26M中的每个问答对包含问题描述和医生或专家的回答。基于该数据集优化的Huatuo-Lite经过多次净化和重写，具有更高的数据维度和质量。

\item LMSYS-Chat-1M\cite{zheng2023lmsys}是一个大规模真实世界LLM对话数据集。它包含了从2023年4月至8月期间在Vicuna演示和Chatbot Arena网站上收集的100万次真实对话，涉及25个最先进的大语言模型。每个样本包括对话ID、模型名称、OpenAI API JSON格式的对话文本、检测到的语言标签和OpenAI审核API标签。该数据集经过脱敏处理以移除个人身份信息，但保留了不安全的对话以供研究人员研究LLM在现实场景中使用的安全相关问题。这个数据集为研究人员提供了宝贵的资源，可用于分析和改进大语言模型在实际应用中的表现。

\item WildChat\cite{Zhao0HC0D24}是一个多语言对话数据集，包含1 039 785个用户与ChatGPT和GPT-4之间的真实对话。这个数据集的独特之处在于它收集了真实场景中用户与AI系统的交互记录，涵盖了各种语言、主题和复杂度的对话。数据收集过程保护了用户隐私，同时保留了对话的自然性和多样性。WildChat反映了用户与AI助手之间的真实交互模式，包括提问方式、澄清需求、处理误解等。这使得该数据集对研究人机对话的动态、提升AI响应的自然度和相关性，以及改进AI助手在实际应用中的表现特别有价值。它可以用于开发更自然、更有适应性的对话系统，改进AI在理解模糊指令、处理多步骤请求和维持长期对话连贯性方面的能力。

\item MMIQC\cite{abs-2401-09003}由从Mathematics Stack Exchange网页中提取的问题-回答对和MetaMathQA中的数据组成。
具体来说，MMIQC从RedPajama数据集中的数学相关网页中提取了1200K问答对进行处理，并要求GPT-4从给定的种子问题中组合新的问题来进行数据扩充。此外，MMIQC提取了MetaMathQA数据中的204k的子集作为补充。MMIQC的构建者基于该数据集微调了一系列模型，其结果显示这些模型在MATH基准上取得了比较大的提升。

\end{itemize}

\subsection{已有数据集扩充}

人工构建数据集和使用真实用户对话作为微调数据集有着各自的优点，同时也需要一定的成本和门槛。
另一种流行的数据集构建方法是利用已有数据集中的数据作为种子数据，通过改写和生成来扩充原有数据，形成规模更大且更全面的数据集。很多实验证明，这种方法可以显著提升模型在微调后的表现。
例如，如表 \ref{MetaMathQA}所示，剑桥大学的研究者通过重新表述、自我验证、正向-反向推理等方式重写GSM8k和Math中的数学问题，并对答案进行扩写，构造了MetaMathQA数据集。这种基于已有数据改写的方法不仅可以快速扩展数据规模，而且能保持数据的多样性和复杂性。基于MetaMathQA\cite{yu2023metamath}数据集微调的模型和基于原数据集微调的模型相比（GSM8k+Math）在数学推理任务上取得了明显的性能提升。

\begin{table}[htb]
\centering
\caption{MetaMathQA数据扩充样例}
\begin{tabular}{p{5cm}p{9cm}}
\toprule
\textbf{原始数据} & \textbf{扩充数据} \\
\midrule

\multirow{3}{5cm}{\textbf{问题:} James buys 5 packs of beef that are 4 pounds each. The price of beef is \$5.50 per pound. How much did he pay?} & \textbf{重新表述:} Question: What is the total amount that James paid when he purchased 5 packs of beef， each weighing 4 pounds， at a price of \$5.50 per pound? \\

&\textbf{自我验证:} James buys x packs of beef that are 4 pounds each. The price of beef is \$5.50 per pound. He paid 110. What is the value of unknown variable x? \\

&\textbf{正向-反向推理:} Question: James buys x packs of beef that are 4 pounds each. The price of beef is \$5.50 per pound. How much did he pay? If we know the answer to the above question is 110， what is the value of unknown variable x? \\

\midrule

\textbf{答案:} He bought 5*4=20 pounds of beef. So he paid 20 * 5.5 = \$110. The answer is: 110 & \textbf{答案扩写:} James buys 5 packs of beef that are 4 pounds each， so he buys a total of 5 * 4 = 20 pounds of beef. The price of beef is \$5.50 per pound， so he pays 20 * $5.50 = $110. The answer is: 110
\\

\bottomrule
\end{tabular}
\label{MetaMathQA}
\end{table}

强语言模型，如ChatGPT，在通过扩充已有数据构建数据集的方法里扮演了关键角色。数据构建者可以提供种子数据和具体的生成要求，依赖强语言模型的生成能力来进行数据改写和扩充。
因此这一数据扩充的过程有时也被看作以构建数据的方式对强语言模型能力的一种学习/拟合。
通过扩充已有数据构建数据集避免了从头开始构建数据集的成本和消耗，节省了大量人力物力资源，同时有效利用了已有数据。然而，尽管语言模型的能力很强，但生成的文本在语言风格和偏好上存在一定的偏向。因此让语言模型生成数据容易导致内容同质、风格类似、信息单一等问题，此外还存在可能出现幻觉的问题。
因此，需要设计合理的流程和规则保证生成数据的多样性、复杂性和正确性，不能盲目的追求构造数据集的规模。许多工作都在这一方面进行了自己的探索，如MetaMathQA通过设计不同的改写目标保证改写数据的多样性。
此外，也有一些数据集通过收集已有数据并选择部分子集构建数据，旨在选择质量高且具有代表性的数据作为微调数据，例如Open-Platypus通过从11个开源数据集中删除重复和类似的问题来构建数据。

表\ref{generated_data3} 列出了一些基于已有数据集扩充构建的数据集和其基本信息，这些数据集的详细介绍如下：

\begin{table*}[htb]
\caption{基于已有数据集扩充构建的数据集}
\centering
\label{generated_data3}
\begin{tabular}{lcc}
\hline
\textbf{数据集名称} & \textbf{样本数量} & \textbf{许可证} \\
\hline
Infinity Instruct & 7M & Apache-2.0 \\
Wizard\_evol\_instruct & 70K-196K & MIT \\
Alpaca\_GPT4\_data\ & 52K & Apache-2.0 \\
Alpaca\_GPT4\_data\_zh & 52K & Apache-2.0 \\
GuanacoDataset & 534.5K & GPL-3.0 \\
CodeFeedback-Filtered-Instruction & 157K & Apache-2.0 \\
OpenMathInstruct-1 & 1.8M & NVIDIA License \\
MetaMathQA & 100K-1M & MIT \\
OIG & 388M & Apache-2.0 \\
Open-Platypus & 24.9K & Apache-2.0 \\
glaive-code-assistant-v3 & 950K & Apache-2.0 \\
code\_bagel\_hermes-2.5 & 3.9M & Apache-2.0 \\
self-oss-instruct-sc2-exec-filter-50k & 50K & ODC-By \\
CodeExercise-Python-27k & 27K & CC-BY-NC-SA-4.0 \\
GSM8K-RFT & 47K &-\\
code\_exercises & 120K & CC-BY-NC-SA-4.0 \\
Evol-Instruct-Code-80k-v1 & 80K & CC-BY-NC-SA-4.0 \\
instructional\_code-search-net-python & 419K & Apache-2.0 \\
DART-Math & 1.1M &MIT \\
\hline
\end{tabular}
\end{table*}

\begin{itemize}

\item Infinity Instruct\cite{BAAI_infinity_instruct}是北京智源人工智能研究院构建的大规模指令数据集。这一数据集旨在补齐模型基础能力（如数学、代码等类型能力），同时提升模型对话能力。Infinity Instruct包含两个主要部分，即模型基础能力相关指令，共计约700万条，主要从现有数据集中选择得出，以及200万条模型对话能力相关指令，利用多种方法自动构建。值得注意的是，Infinity Instruct对话能力相关指令包含完成该指令所需何种能力与知识的标签。这有助于使用者了解数据分布，按需选择不同类型的指令。

\item Wizard\_evol\_instruct\cite{bib68}数据集由微软等机构发布，包含196 000个实例。该数据集通过将现有的指令数据进行演变，以生成更加复杂和多样化的数据集，它包含143 000个由Alpaca和ShareGPT数据混合演变而来的样本。Wizard\_evol\_instruct专注于提升指令的多样性和复杂性，以改进大语言模型的训练效果，特别针对于代码生成领域。

\item Alpaca\_GPT4\_data\cite{vicgalle_alpaca_gpt4}是一个由Microsoft Research发布的高质量英文指令数据集，包含52 000条由GPT-4生成的指令-回答对。该数据集使用Apache-2.0许可证，基于Alpaca项目的提示词生成，涵盖多种指令类型。数据以JSON格式存储，每条数据包含唯一的指令、可选输入和GPT-4生成的输出。这个数据集主要用于微调大语言模型，提升其指令跟随能力。人工评估显示，使用此数据集训练的模型在“有帮助性”方面明显优于使用GPT-3数据训练的模型，且在多个评估指标上接近GPT-4的表现，展现了其在构建先进指令跟随型语言模型方面的潜力。

\item Alpaca\_GPT4\_data\_zh\cite{peng2023gpt4llm}是Alpaca\_GPT4\_data的中文版本，同样由Microsoft Research发布，包含52 000条指令-回答对。这个数据集的生成过程首先使用ChatGPT将Alpaca项目的英文提示词翻译成中文，然后用这些中文提示词指导GPT-4生成回答。它与英文版共享相同的许可证和数据格式，主要用于提升中文大语言模型的指令理解和执行能力。作为GPT-4-LLM项目的重要组成部分，这个数据集为研究人员提供了宝贵的资源，用于构建和优化基于监督学习和强化学习的中文指令跟随型大语言模型。

\item GuanacoDataset\cite{JosephusCheung_guanaco_dataset}是由 是由JosephusCheung发布的一个多任务、多语言的数据集，包含 534530 个实例。该数据集由Alpaca初始 52K数据集基础上扩展而来，旨在增强模型在多语言环境中处理各种语言任务的能力。GuanacoDataset包含中文、英文、日文等多种语言，并涵盖文本生成、问答等任务。数据集提供了丰富的训练实例，旨在提升模型理解和执行多样化任务指令的能力，从而为多任务学习和跨语言研究提供强大的支持。

\item CodeFeedback-Filtered-Instruction\cite{ZhengZSLLFCY24}是一个经过筛选的代码指令查询集合，提取自四个主要的开源代码指令调优数据集，包括Magicoder-OSS-Instruct、ShareGPT的 Python代码子集、Magicoder-Evol-Instruct和 Evol-Instruct-Code。最初，从这些数据集中汇总了 28.7 万个查询。为了筛选出最复杂和信息量最大的指令，采用了严格的筛选过程。使用开源聊天模型Qwen-72B-Chat进行选择性过滤，对代码查询及其对应的响应进行评估，并赋予复杂性评分（从 1 到 5），仅保留评分为 4 或 5 的查询。这一细致的筛选过程最终得到了一套 15.6 万条高质量的单轮代码指令。在数据处理的后续步骤中，除了单轮打包，我们保留了所有响应，以便为用户提供更多的使用选项。该数据集遵循Apache-2.0 许可证，包含 15.7 万行数据。

\item OpenMathInstruct-1\cite{bib411}是由NVIDIA发布的一个大规模数学指令调优数据集，包含180万个问题-解决方案对。这个数据集是使用Mixtral-8x7B模型基于GSM8K和MATH数据集的训练子集生成的。它的主要特点是结合了文本推理和可执行的Python代码块来解决数学问题，为数学领域的机器学习研究提供了丰富的资源。数据集采用NVIDIA许可证，分为训练集和验证集，完整覆盖了GSM8K和MATH的全部训练集。OpenMathInstruct-1专注于数学领域，旨在提高模型在数学问题解决和推理方面的能力。该数据集的发布为数学教育、自动问题解答系统的开发，以及人工智能在数学领域的应用提供了宝贵的训练资源。这个数据集的规模和质量使其成为数学领域机器学习研究的重要资源。

\item MetaMathQA\cite{yu2023metamath}是由剑桥大学等机构发布的一个专注于数学推理的大规模数据集。该数据集基于GSM8K和MATH数据集的训练集通过问题重写和增强生成，包含10万到100万个实例。MetaMathQA的独特之处在于它采用了多角度问题重写的方法来扩充原始数学问题，旨在提高语言模型在数学推理方面的能力。数据集中的每个条目都包含一个原始问题（来自GSM8K或MATH训练集）及其重写版本。这个数据集是为了解决开源大语言模型（如LLaMA-2）在复杂数学问题解决方面的不足而创建的。研究人员使用MetaMathQA对LLaMA-2模型进行微调，创建了MetaMath模型，在GSM8K和MATH基准测试中取得了显著的性能提升。MetaMathQA采用MIT许可证，为研究人员和开发者提供了宝贵的资源，用于提升语言模型在数学推理方面的能力。该数据集的发布，连同预训练的MetaMath模型和训练代码，为数学问题求解和人工智能教育领域的研究提供了重要支持。这个数据集的创新性和规模使其成为研究数学推理、问题重写技术以及大语言模型在特定领域应用方面的重要资源。

\item OIG （Open Instruction Generalist）\cite{ontocord_oig_moderation}数据集是由LAION发布的大规模英语指令数据集，包含近388万个实例。这个数据集旨在创建一个中等质量的大型指令数据集和一个较小的高质量指令数据集（OIG-small-chip2）。数据以JSONL格式存储，包含人机对话形式的指令和回复。OIG的目标是提供足够的多样化指令数据来预训练语言模型，然后在较小的高质量数据集上进行微调，从而将预训练模型转化为指令跟随模型。该项目计划不断扩大数据集规模，最终目标是创建1万亿个词元的多样化指令数据。OIG采用Apache-2.0许可证，涵盖多种指令类型，为研究人员提供了丰富的资源来开发和改进指令跟随型语言模型。

\item Open-Platypus\cite{garage-bAInd_open_platypus}是由波士顿大学发布的一个英语指令数据集，包含约25 000个实例。这是一个精心策划的数据集，主要聚焦于提升大语言模型在STEM和逻辑推理方面的能力。数据来源于11个开源数据集，经过关键词搜索和句子相似度过滤，去除了重复和相似度高的问题。其中约90\%的问题是人工编写的，确保了数据质量。这个数据集的独特之处在于它能够在最小化微调时间和成本的情况下，显著提升模型性能。Open-Platypus被用于训练Platypus系列模型，这些模型在多项LLM评估指标上取得了优秀成绩。该数据集为研究人员提供了宝贵的资源，用于探索如何高效地提升大语言模型的逻辑推理能力。

\item glaive-code-assistant\cite{glaiveai_code_assistant}是由glaiveai公司发布的一个大规模关于代码生成问题与解决方案数据集。该数据集包含约 100 万个通过Glaive平台的合成数据生成平台生成的代码问题和解决方案。共有v1、v2 和v3 三个版本。Glaive-code-assistant-v3 数据集共有 95 万行数据。该数据集采用Apache-2.0 许可证发布，为研究人员和开发者提供了丰富的资源，以探索和优化代码生成与解决方案的各种应用。

\item code\_bagel\_hermes-2.5\cite{OpenHermes2.5} 是由Replete-AI发布的一个综合性数据集，结合了code\_bagel和 Open-Hermes-2.5 数据集。该数据集遵循Apache-2.0 许可证，提供了大约 90 万行高质量的非代码指令数据和 300 万行高质量的代码指令数据，总计 390 万行数据。每行数据最多包含 10 000 个标记，并支持超过 100 种编程语言。这一数据集采用了sharegpt结构，将数据以字典列表的形式组织，每个字典包含一个对话的多个回合，其中每个回合包括角色和文本内容。Replete-AI/code\_bagel\_hermes-2.5 旨在为编程模型的微调提供丰富的数据支持，特别适用于开发和优化编程相关的模型。数据集的广泛语言支持和大规模内容使其成为提升编程任务性能的理想选择。

\item self-oss-instruct-sc2-exec-filter-50k\cite{bigcode_self_oss_instruct_sc2_exec_filter_50k}是用于训练StarCoder2-15B-Instruct-v0.1 的数据集，主要用于Python代码生成任务。该数据集是通过一个完全自我对齐的开源管道生成的，旨在提供一个透明的、无依赖于人工标注或从大型专有模型中提取的数据的训练数据集。

\item CodeExercise-Python-27k\cite{codefuse_ai_codeexercise_python_27k}是由codefuse创建的一个数据集，包含 2.7 万道Python编程练习题（英文），覆盖了基础语法与数据结构、算法应用、数据库查询、机器学习等数百个Python相关知识点。该数据集借助老师模型和Camel生成，旨在提供广泛的编程练习资源。由于数据集未经严格校验，题目或答案可能存在错误或语义重复，使用时需谨慎。数据集采用CC-BY-NC-SA-4.0 许可证发布

\item code\_exercises\cite{jinaai_code_exercises}数据集由Jina.ai开发，包含约 12 万个由ChatGPT 3.5 生成的Python代码练习（总计约 1.2 亿个token）。该数据集旨在将ChatGPT 3.5 对Python编程任务的知识提炼到其他（可能更小的）模型中，生成的练习遵循Human Eval基准的格式，每个练习包括一个带有描述性文档字符串的Python函数签名和一个解决方案。此数据集采用CC-BY-NC-SA-4.0 许可证发布，并且通过系统提示确保了数据的多样性，以提高模型在特定任务上的表现。

\item GSM8K-RFT\cite{li2024query}通过拒绝采样来生成和收集正确的推理路径，依此来扩充数学推理数据集GSM8K。
RFT方法首先利用预训练模型生成大量候选输出，然后通过特定的筛选机制（如人工评审或自动评分）从中挑选出高质量的样本。通过生成包含更多独特推理路径的增强样本，GSM8K-RFT相比于原始的GSM8K数据集，能够更多地提升大型语言模型（LLMs）的数学推理能力。此外，实验表明这种提升对于性能较差的LLMs可能更为明显。

\item Evol-Instruct-Code-80k-v1\cite{luo2023wizardcoder}是由Nick Roshdieh机构发布的一个专注于代码指令的数据集，包含约80 000个实例。该数据集由领域专家精心策划，涵盖了Python、Java、C++、JavaScript等多种编程语言的指令示例。它旨在提供一个丰富的资源，用于训练和评估代码生成、自然语言处理和程序合成等机器学习模型。数据集以CSV格式提供，每个条目包含代码指令和相应的输出，涉及广泛的编程概念、技术和问题解决方法。Evol-Instruct-Code-80k-v1采用cc-by-nc-sa-4.0通用许可，允许自由使用和修改，为研究人员和开发者提供了宝贵的工具，以推进代码理解和生成领域的研究，并有助于改进自动编程工具和编程教育资源。

\item instructional\_code-search-net-python\cite{nan_do_instructional_code_search_net_python}是一个专注于Python编程的指令型数据集，于2023年5月创建。该数据集旨在提高大型语言模型（LLMs）的编码能力，采用Apache 2.0许可证。这个数据集包含两种主要任务类型：给定一段代码，生成描述其功能的文本。给定一段功能描述，生成相应的Python代码。该数据集基于code-search-net数据集的摘要版本进行构建，使用模板和自然语言处理技术生成类人的指令和响应。数据集的创建过程重点确保了没有重复或无意义的摘要。通过提供大量的代码-描述对，instructional\_code-search-net-python为研究人员和开发者提供了宝贵的资源，有助于改进AI系统在Python编程领域的能力，包括代码生成、代码理解和自然语言到代码的转换等任务。

\item DART-Math\cite{abs-2407-13690}基于GSM8k和MATH进行数据合成，通过对困难问题进行更多的采样，合成与已有方法相比更重视困难问题的数学数据集。具体来说，构建者通过开源模型在一个数学问题上的失败了来评估该问题的难度，并基于困难的问题构造更多的训练样本。

\end{itemize}

\subsection{语言模型直接生成}
语言模型除了可以帮助扩充已有数据集外，有时还被用于直接生成数据。如加利福尼亚大学通过让ChatGPT进行角色场景扮演，并进行自我对话，通过这种方式构建了多轮对话数据集：Baize。同利用语言模型进行数据扩充一样，利用语言模型直接生成数据也存在内容同质、风格类似、信息单一、幻觉等问题。因此需要进行针对性的设计保证数据的多样性和质量。即使是这样，相比于使用前几种方式构造的数据，利用语言模型直接生成的数据往往被看作多样性不足的数据源。其可以看作强语言模型蕴含知识的数据形式的表示，作为补充数据使用，一般不作为微调模型的主要数据来源。

表\ref{generated_data4} 列出了一些使用语言模型直接生成的数据集和其基本信息，这些数据集的详细介绍如下：

\begin{table*}[htb]
\caption{使用语言模型直接生成的数据集}
\centering
\label{generated_data4}
\begin{tabular}{lcc}
\hline
\textbf{数据集名称} & \textbf{样本数量} & \textbf{许可证} \\
\hline
Alpaca\_data & 52K & Apache-2.0 \\
BELLE系列数据集 & 0.5M-3.5M & GPL-3.0 \\
CAMEL & 1.66M & CC-BY-NC-4.0 \\
MOSS系列数据集 & 1M & CC-BY-NC-4.0 \\
Chatgpt\_corpus & 3.27M & GPL-3.0 \\
Self-Instruct & 52.4K & Apache-2.0 \\
UltraChat & 1.47M & CC-BY-NC-4.0 \\
RedGPT & 89K & Apache-2.0 \\
generated\_chat\_0.4M & 400K & GPL-3.0 \\
InstructIE & 371.7K & MIT \\
orca-math-word-problems-200k & 200K & MIT \\
OpenHermes-2.5 & 1M &-\\
python-code-dataset-500K & 560K & MIT \\
Code\_Alpaca\_20K & 20K & Apache-2.0 \\
HC3 & 37K & CC-BY-SA-4.0\\
Camel-ai code & 50K & CC-BY-NC-4.0 \\
Camel-ai math & 50K & CC-BY-NC-4.0 \\

\hline
\end{tabular}
\end{table*}

\begin{itemize}
\item Alpaca\_data\cite{taori2023stanford}是斯坦福大学在2023年3月发布的数据集，使用OpenAI的text-davinci-003模型生成了52 000个高质量的指令-输入-输出三元组。这个数据集涵盖了广泛的任务类型，包括创意写作、代码生成、问答、分析和推理等。Alpaca\_data的创新在于使用了自我指导（self-instruct）方法，通过大语言模型自动生成多样化的指令和相应的输出。这种方法显著降低了创建大规模指令数据集的成本和时间。该数据集在低资源环境下微调语言模型方面表现出色，特别适合用于研究如何使用较少的数据和计算资源来改进大型语言模型的性能。它在开发个性化AI助手、特定领域的语言模型和提升模型的指令跟随能力方面有重要应用。

\item BELLE系列数据集BELLE（Better Environment for Learning LLM Efficiently）\cite{BELLE,ji2023better}是一系列由大型语言模型（如ChatGPT或GPT-3）生成的中文指令数据集，规模从50万到350万不等。这些数据集涵盖了多种任务类型和领域，包括对话、问答、文本分类、摘要生成等。BELLE的独特之处在于它专注于中文语境，通过多轮优化和人工筛选确保了数据质量。数据集的创建过程包括指令模板设计、任务特定数据生成和人工质量控制。BELLE旨在提供大规模、高质量的中文指令数据，用于训练和微调中文语言模型。这些数据集对提升模型在中文环境下的表现特别有价值，可应用于开发中文AI助手、智能客服系统、自动写作工具等。

\item CAMEL（Communicative Agents for ``Mind" Exploration of Large Scale Language Model Society）\cite{li2023camel}是一个包含1 659 328个对话实例的数据集，由两个GPT-3.5-Turbo代理通过相互对话生成。这个数据集模拟了不同角色（如科学家、工程师、艺术家等）之间的专业对话，涵盖多种语言和编程语言。CAMEL的创新之处在于其独特的数据生成方法：通过定义不同的角色和任务目标，让AI代理自主进行对话，从而产生丰富多样的交互数据。这种方法不仅提高了数据的多样性，还探索了AI系统在复杂社交情境中的行为。CAMEL对研究AI代理之间的交互、协作问题解决和角色扮演对话生成特别有价值。它可应用于开发更高级的对话系统、虚拟协作环境和社交AI等领域。

\item MOSS系列数据集\cite{Sun2024MOSS}是由复旦大学开发的一系列大规模多轮对话数据集，旨在训练和改进大型语言模型。这些数据集涵盖了多个版本和应用场景: moss-002-sft-data数据集包含约116万条由text-davinci-003生成的中英文对话，涉及模型的有用性、忠实性和无害性。moss-003-sft-data数据集基于moss-002-sft-data的用户反馈和GPT-3.5-turbo生成，包含约107万条更贴近真实用户意图的对话，具有更细致的标注和更长的对话轮次。
moss-003-sft-plugin-data插件数据集包含约30万条支持搜索引擎、文生图等插件的多轮对话数据，拓展了模型的功能范围。
moss-003-pm-data偏好数据集基于额外的对话上下文和moss-003-sft-data的输出构建，用于进一步优化模型表现。
这些数据集涵盖了多语言（中英文）和多种任务类型，为研究人员提供了丰富的资源来探索和改进大型语言模型在各种场景下的性能。

\item Chatgpt\_corpus\cite{plexpt_chatgpt_corpus}是一个大规模中文对话数据集，包含3 270 000个由GPT-3.5-Turbo生成的各种主题和场景的对话。数据集涵盖了日常交谈、专业咨询、问题解答、创意写作等多种类型的对话。Chatgpt\_corpus的特点是利用先进的语言模型生成大量高质量、多样化的中文对话数据。这种方法克服了人工创建大规模对话数据集的困难，同时保证了内容的丰富性和语言的自然度。该数据集为训练中文聊天机器人和对话系统提供了丰富的资源，特别适合用于开发开放域对话系统、情感分析工具、自动客服系统等。它在提升模型的中文语言理解、上下文把握和回应生成能力方面具有重要作用。

\item Self-Instruct\cite{WangKMLSKH23}是华盛顿大学等机构开发的数据集，使用GPT-3自动生成52 445个指令-输出对。Self-Instruct采用了创新的自我指导方法来生成数据：首先使用少量人工编写的种子任务，然后让语言模型自主生成新的指令和相应的输出。这种方法不仅大大减少了人工标注的工作量，还能产生多样化和创新性的指令。数据集涵盖了广泛的任务类型，从简单的问答到复杂的推理和创作任务。Self-Instruct的主要优势在于它能够生成大量多样化的指令，有助于提高模型的泛化能力和适应新任务的能力。这个数据集特别适合用于研究如何提升语言模型的指令跟随能力、探索模型的零样本和少样本学习能力，以及开发更灵活、适应性更强的AI系统。

\item UltraChat\cite{ding2023enhancing}是清华大学发布的大规模英文对话数据集，包含1 468 352个通过两个ChatGPT代理之间的对话生成的实例。UltraChat覆盖了广泛的主题和场景，包括日常对话、学术讨论、创意写作、问题解决等。数据集的创建采用了创新的方法：通过精心设计的提示词引导两个AI代理进行多轮对话，模拟人类之间的自然交互。这种方法不仅能生成大量高质量的对话数据，还能探索AI系统在复杂对话情境中的表现。UltraChat对训练开放域对话系统特别有用，可以显著提升模型的上下文理解、多轮对话管理和生成连贯回应的能力。它在开发高级聊天机器人、虚拟助手、对话式推荐系统等方面有广泛的应用前景。

\item RedGPT\cite{redgpt}数据集是一个自动生成的中文多轮对话数据集，旨在提升大语言模型的事实正确性。它采用“Reference-Enlightened-Dialogue GPT”（RedGPT）方法，基于高质量事实型文档自动生成对话。第一批公开数据包含5万条事实型知识对话和3.9万条编程相关对话，涵盖了人物、科技、医疗、法律、艺术等多个领域。这个数据集通过精选参考资料和精心设计的提示词确保生成对话的质量和多样性，为研究人员提供了一种低成本、可扩展的方法来生成大规模、高质量的事实型对话数据，有助于提升语言模型在事实性问答方面的表现。

\item generated\_chat\_0.4M\cite{bellegroup_generated_chat_0_4M}是由BELLE项目发布的一个大规模中文个性化角色对话数据集。它包含约40万条由ChatGPT生成的对话实例，每个实例都包含角色介绍和相应的对话内容。这个数据集旨在为中文大语言模型提供丰富的角色扮演训练资源，有助于提升模型在多样化场景下的对话能力。值得注意的是，由于数据是由AI自动生成的，未经过严格的人工验证，可能存在一些事实性错误或不准确之处。研究者在使用时需要注意这一点，并根据需要进行额外的筛选或验证。这个数据集采用GPL-3.0许可证，为研究人员提供了探索个性化对话生成和角色扮演任务的宝贵资源。

\item InstructIE\cite{abs-2305-11527}是一个专注于信息抽取（IE）任务的指令数据集，包含371 700个从百度百科和维基百科收集的英文和中文数据。数据集涵盖了多种IE任务，如命名实体识别、关系抽取、事件抽取等。每个实例都包含详细的任务指令、输入文本和预期输出，使模型能够学习如何根据不同的指令执行各种IE任务。InstructIE的创新之处在于将IE任务框架化为指令跟随问题，有助于提高语言模型在特定领域信息抽取任务上的灵活性和性能。这个数据集对开发可适应不同IE场景的通用模型特别有价值，可应用于自动化信息提取、知识库构建和智能文档处理等领域。

\item orca-math-word-problems-200k\cite{mitra2024orcamath}是由Microsoft发布的一个专注于小学数学应用题的数据集，包含约20万个问题-答案对。这个数据集是使用Azure GPT-4 Turbo基于一个种子问题集扩展生成的，所有答案也都由Azure GPT-4 Turbo生成。数据集采用MIT许可证，主要用于增强语言模型的数学能力，为数学问题解决提供坚实的基础。每个数据条目包含一个问题及其对应的答案。其主要目的是为语言模型在数学领域的研究和应用提供训练资源。这个数据集的规模和专注性使其成为研究语言模型在数学问题解决能力方面的重要资源。它为探索人工智能在基础数学教育和问题解答方面的潜力提供了宝贵的数据支持。

\item OpenHermes-2.5\cite{OpenHermes2.5}是由Teknium发布的一个大规模指令和对话数据集，包含100万个样本。这个数据集是Open Hermes 1的延续和扩展，在规模、多样性和质量上都有显著提升。它融合了多个开源数据集和定制创建的合成数据集，主要由合成生成的指令和聊天样本组成。OpenHermes-2.5是Open Hermes 2.5和Nous Hermes 2系列模型的基础训练数据，旨在推动大型语言模型（LLM）的发展，使其在各种任务中达到最先进的水平。这个数据集的特点是大规模、高质量、多样性强，主要由人工智能生成的高质量指令和对话组成。它为研究人员和开发者提供了宝贵的资源，用于训练和改进大型语言模型，特别是在指令跟随和对话生成方面，对推动开源AI模型的发展和提高其性能做出了重要贡献。

\item python-code-dataset-500k\cite{jtatman_python_code_dataset_500k}是一个包含约56万条数据的Python代码相关数据集，通过从GitHub代码中提取和重新格式化而成，采用MIT许可证。该数据集的结构包含三个主要字段：instruction（用户指令）、system（系统指令）和output（输出的代码内容）。

\item Code\_Alpaca\_20K\cite{sahil2801_codealpaca_20k}是由Sahil Chaudhary发布的一个用于指令驱动代码生成的数据集。该数据集包含20 000个独特的指令-输出对，专注于代码生成、编辑和优化任务。数据是使用OpenAI的text-davinci-003模型生成的，采用Apache-2.0许可证。Code\_Alpaca\_20K的生成过程借鉴了Stanford Alpaca项目，但进行了针对性修改，使其更专注于代码相关任务。这种方法不仅产生了高质量的数据集，还大大降低了生成成本（不到200美元）。数据集涵盖多种编程语言，主要以英语为主，为代码生成、自然语言处理在编程领域的应用提供了宝贵的资源。

\item HC3（Human ChatGPT Comparison Corpus）\cite{guo-etal-2023-hc3}是由SimpleAI团队发布的一个大规模数据集，包含37175个问答对实例，收集了人类专家和ChatGPT针对相同问题的回答，涵盖开放域、金融、医疗、法律、心理学等多个领域。其目的在于研究ChatGPT与人类专家回答的特征差异，并开发ChatGPT文本检测器。论文基于该数据集进行了全面的人工评估和语言学分析，揭示了许多有趣的发现，并基于HC3开发了三种不同的ChatGPT检测系统。该数据集、代码和模型都公开发布在GitHub上，为研究大语言模型与人类专家的差距及探讨LLM未来发展方向提供了宝贵资源。

\item Camel-ai math\cite{camel_ai_math}数据集是由沙特阿拉伯阿卜杜拉国王科技大学（KAUST）研究人员构建的大规模数学问题解决数据集。这一数据集包含50K问题-解答对，全部由GPT-4模型生成。CAMEL Math涵盖了25个数学主题，每个主题下设25个子主题，每个“主题-子主题”组合包含80个问题。这种结构设计旨在全面覆盖各种数学领域和难度水平。

\item Camel-ai code\cite{camel_ai_code}是由KAUST研究人员构建的大规模对话数据集。这一数据集由50K次两个GPT-3.5-turbo代理之间的对话组成，模拟了专精于特定编程语言的程序员与来自特定领域的人员之间的交互。CAMEL涵盖了20种编程语言和50个领域，每种语言和领域组合包含50个任务

\end{itemize}

\subsection{传统数据集改写}

在自然语言处理（NLP）的发展历程可以说是构建任务解决任务的一个过程。从最基础的分词、句法分析，到文本分类、机器翻译、命名实体识别，再到机器阅读理解，常识问答，NLP能够解决的任务越来越多也越来越复杂。
其中每个任务的研究和发展都依赖其相关的数据集的构建，这些数据集为NLP的发展和应用提供了重要支持。例如，自2016年斯坦福大学发布的SQuAD（Stanford Question Answering Dataset）\cite{RajpurkarJL18,RajpurkarZLL16}机器阅读理解数据集到2024年，机器阅读理解研究领域新增了上百个数据集，这些数据集覆盖了文学、科技、医疗和法律等多个领域。
虽然大语言模型的微调数据集可能在形式和侧重与一些传统的NLP数据集有所不同，但是这些传统数据集的构建思想和数据标注中涵盖的语言知识也能帮助大语言模型学习如何更好地解决用户的问题。
因此，一种微调数据构建方式是把传统NLP任务的数据集转换为指令跟随/对话的形式，这样便可以应用在大语言模型的微调中。
例如，Allen AI发布的NATURAL INSTRUCTIONS\cite{naturalinstructions, supernaturalinstructions}将各种传统NLP任务表述为自然语言指令，数据集涵盖了文本分类、问答、摘要生成、文本生成等多种任务类型，每个任务都配有详细的任务描述，包含任务定义，注意事项，任务样例等。通常基于传统数据集改写可以以较低成本获得大量数据，并且由于很多传统数据集是人工标注的并且有着较为严格的质量检测过程，所构造数据质量相对较高。然而，传统数据集的数量虽然庞大，但也是有限的。并且并不是所有的数据集都有着较高质量和丰富的语言知识。此外，很多传统任务，例如情感分类任务对于目前的预训练模型可能过于简单，导致其对模型的改善较为有限。

表\ref{generated_data5} 列出了一些基于传统数据集改写的数据集和其基本信息，这些数据集的详细介绍如下：

\begin{table*}[htb]
\caption{基于传统数据集改写的数据集}
\centering
\label{generated_data5}
\begin{tabular}{lcc}
\hline
\textbf{数据集名称} & \textbf{实例/子数据集数量} & \textbf{许可证} \\
\hline
CrossFit & 269个数据集 &-\\
DialogStudio & 87个数据集 & Apache-2.0 \\
Dynosaur & 801.9K实例 & Apache-2.0 \\
Flan系列 & 62-1836个数据集 & Apache-2.0 \\
NATURAL INSTRUCTIONS & 61个数据集 & Apache-2.0 \\
SUPER-NATURAL INSTRUCTIONS & 1616个数据集 & Apache-2.0 \\
OPT-IML Bench & 2000个数据集 & MIT \\
PromptSource & 176个数据集 & Apache-2.0 \\
T0 & 62个数据集 & Apache-2.0 \\
IEPile & 33个数据集 & CC-BY-NC-SA-4.0 \\
firefly-train-1.1M & 1.15M实例 & Apache License 2.0 \\
xP3 & 82个数据集 & Apache-2.0 \\
\hline
\end{tabular}
\end{table*}

\begin{itemize}

\item CrossFit\cite{ye-etal-2021-crossfit}数据集由南加州大学发布，包含269个经过改进的自然语言处理（NLP）数据集。该数据集用于研究在跨任务中实现少样本学习的能力。CrossFit旨在构建能够在不同NLP任务上进行少样本学习的模型，例如，研究训练非分类任务的模型是否能在分类任务上表现出色，或训练非机器阅读理解问答任务的模型是否能在机器阅读理解问答任务上成为优秀的少样本学习者。

\item DialogStudio\cite{ZhangQLHMLYWSX24}数据集由Salesforce AI等机构发布，包含87个对话数据集。该数据集通过收集和改进各种自然语言处理（NLP）数据集形成，并遵循Apache-2.0许可发布。DialogStudio统一了每个数据集，同时保留了其原始信息，这有助于支持对单个数据集的研究以及大型语言模型（LLM）的训练。该数据集旨在为对话式人工智能的研究提供广泛而多样的数据资源。

\item Dynosaur\cite{bib76}数据集由UCLA等机构发布，包含801 900个实例，遵循Apache-2.0许可发布。该数据集通过收集和改进各种NLP数据集形成。Dynosaur提出了一种动态增长范式，用于指令微调数据的策划。基于NLP数据集的元数据，生成适用于各种NLP数据集的多任务指令，并确定相关的数据字段，以构建指令微调数据。Dynosaur数据集具有较低的生成成本，生成80万条指令微调数据的成本不到12美元。其指令微调数据质量高，比Self-Instruct的有效性更高，在Super-NI上的下游性能优于Alpaca和Instruction GPT-4。此外，Dynosaur具有动态增长能力，能够持续生成新的数据集指令，为改进指令跟随模型提供了一个不断扩展的资源。研究表明，在Super-NI任务上，使用Dynosaur数据微调的模型性能优于使用Alpaca、Instruction GPT-4和Dolly数据微调的模型。

\item Flan系列（Fine-tuned LAnguage Net）\cite{iclr/WeiBZGYLDDL22}是谷歌研究院开发的一系列数据集集合，整合了62至1836个现有的NLP数据集，并进行了改进和标准化。Flan系列的创新在于它采用了“指令微调”的方法，将各种NLP任务重新构造为自然语言指令的形式。这种方法显著提高了模型在多样化任务上的性能，特别是在零样本和少样本场景下。Flan数据集覆盖了广泛的任务类型，包括分类、问答、摘要生成、文本生成等，并支持多种语言。它的主要优势在于能够训练出更加通用和灵活的语言模型，这些模型能够理解和执行各种自然语言指令。Flan系列在提升大型语言模型的泛化能力、跨任务迁移学习和指令跟随能力方面发挥了重要作用，为开发更智能、更通用的AI系统奠定了基础。

\item NATURAL INSTRUCTIONS\cite{naturalinstructions,supernaturalinstructions}是Allen AI研究所等机构开发的数据集合，包含61个不同的NLP任务数据集。NATURAL INSTRUCTIONS的核心理念是将各种NLP任务表述为自然语言指令，每个任务都配有详细的任务描述、正面和负面示例、以及相关的输入-输出对。这种方法旨在提高模型理解和执行多样化任务指令的能力，使模型能够更好地泛化到新任务。数据集涵盖了文本分类、问答、摘要生成、文本生成等多种任务类型。NATURAL INSTRUCTIONS的创新之处在于它强调了任务指令的重要性，使模型能够“读懂”任务需求，而不仅仅是记忆特定的模式。这个数据集对研究如何构建更灵活、更具适应性的NLP模型有着重要意义.

\item SUPER-NATURAL INSTRUCTIONS\cite{WangMAKMNADASPK22}是NATURAL INSTRUCTIONS的扩展版本，包含1616个不同的NLP任务数据集。相比于原版本，SUPER-NATURAL INSTRUCTIONS大幅增加了任务的多样性和复杂性，涵盖了更多的语言和任务类型，如机器翻译、摘要生成和对话生成等。这些数据集为研究人员提供了更广泛的测试平台，有助于深入理解和改进大规模多任务学习和跨语言迁移的能力。其全面的任务覆盖和多样化的数据特性使其成为评估和优化模型性能的重要资源。

\begin{table}[t!]
\centering
\caption{微调数据集构建方式总结}
\resizebox{0.9\textwidth}{!}{
\begin{tabular}{lll}
\toprule
\textbf{构建方式} & \textbf{优点} & \textbf{缺点} \\
\midrule
\textbf{人工撰写} & \begin{tabular}[c]{@{}m{5cm}@{}}-高质量和准确性\\-数据安全性与无害性可控\\-指令和回答可控性强\end{tabular} & \begin{tabular}[c]{@{}m{5cm}@{}}-成本较高\\-多样性有限，难以覆盖各种实际应用场景\\ \end{tabular} \\
\midrule
\textbf{真实用户数据} & \begin{tabular}[c]{@{}m{5cm}@{}}-反映真实场景中的用户需求和交互模式\\-自然性和多样性高 \end{tabular} & \begin{tabular}[c]{@{}m{5cm}@{}}-可能包含有害信息、噪声和无效对话\\-需确保用户隐私得到保护\end{tabular} \\
\midrule
\textbf{基于已有数据集扩充} & \begin{tabular}[c]{@{}m{5cm}@{}}-增加已有数据的复杂性和多样性 \\-快速扩展数据规模，节省成本和资源 \\ \end{tabular} & \begin{tabular}[c]{@{}m{5cm}@{}}-生成数据可能存在同质化问题 \\-需设计合理流程以保证多样性和数据质量 \end{tabular} \\
\midrule
\textbf{语言模型直接生成} & \begin{tabular}[c]{@{}m{5cm}@{}}-用非常低的成本生成大量数据 \end{tabular} & \begin{tabular}[c]{@{}m{5cm}@{}}-数据多样性不足 \\-内容存在同质化和幻觉问题 \end{tabular} \\
\midrule
\textbf{传统数据集改写} & \begin{tabular}[c]{@{}m{5cm}@{}}-利用了传统数据集的构建思想和蕴含的语言知识 \end{tabular} & \begin{tabular}[c]{@{}m{5cm}@{}}-任务难度不足 \\-高质量数据集数量有限 \end{tabular} \\
\bottomrule
\end{tabular}
}
\label{sft_summary}
\end{table}

\item OPT-IML Bench\cite{iyer2022opt}是由Meta AI发布的一个大规模自然语言处理数据集集合。它包含了2000个NLP任务数据集，涵盖93种不同的任务类型。这个数据集集合整合并筛选了八个大型数据仓库的内容，包括CrossFit、UnifiedSKG和PromptSource等。OPT-IML Bench的主要目的是用于研究指令微调过程中的各种决策对下游任务性能的影响。这个多语言、多任务类型的数据集采用MIT许可证，为研究人员提供了丰富的资源来探索和优化大语言模型在各种NLP任务上的表现。

\item PromptSource\cite{BachSYWRNSKBFAD22}是由包括布朗大学在内的机构开发的一个自然语言处理数据集集合和提示工具系统。这个数据集包含176个NLP任务数据集，涵盖13种不同的任务类型。PromptSource的主要特点是它提供了一个多样化的提示集合，为指令微调等研究领域提供了丰富的资源。该系统允许用户创建、共享和使用自然语言提示，这些提示可以将数据集中的示例映射到自然语言输入和目标输出。PromptSource采用Apache-2.0许可证，包含了超过2000个针对约170个数据集的提示，为研究人员提供了一个协作开发和优化语言模型提示的平台。

\item T0\cite{sanh2021multitask}是Hugging Face等机构开发的数据集集合，包含62个专门用于训练和评估零样本学习能力的数据集。T0数据集通过整合来自不同来源的数据集并重新格式化为提示-完成对，提供了一个综合性的零样本学习测试平台。这些数据集覆盖了多个任务领域，包括文本生成、问答和分类任务，旨在提升模型在没有明确训练样本的情况下的表现。通过利用这些数据集，研究人员可以评估模型在处理新颖任务和语言时的适应能力和泛化性能。（规模：62个数据集，许可证：Apache-2.0）

\item IEPile\cite{abs-2402-14710}是由浙江大学等机构发布的一个大规模信息抽取（IE）数据集。这个数据集整合了26个英语和7个中文IE数据集，涵盖了通用、医疗、金融等多个领域，包含约3.2亿个词元。
IEPile的独特之处在于其采用了``基于模式的批量指令生成方法“来创建高质量的IE微调数据。这种方法解决了传统IE数据集在构建指令时可能出现的问题，如训练和评估时模式查询数量不一致，以及指令中模式区分不充分等。
研究者引入了硬负样本模式和批量指令生成两种策略来优化数据集。这些方法有助于提高模型对语义相近模式的区分能力，同时减少训练实例数量而不影响模型性能。
基于IEPile数据集，研究者对Baichuan2-13B-Chat和LLaMA2-13B-Chat模型进行了微调，在全监督训练集上表现出色，并在零样本信息抽取任务中取得了显著进步。
IEPile采用CC-BY-NC-SA-4.0许可证，为研究人员提供了丰富的资源来探索和改进信息抽取任务的性能。

\item firefly-train-1.1M\cite{yeungnlp_firefly_train_1_1M}是Firefly项目的一部分，由YeungNLP发布，包含约165万个实例，涵盖23种常见的中文NLP任务。它特别注重中华文化相关内容，如对联、诗歌和文言文翻译等。数据集采用人工编写的多种指令模板，确保了高质量和多样性，旨在为中文语言模型的指令微调提供全面的训练资源，特别适合需要处理中文文本和理解中国文化的AI模型。

\item xP3\cite{muennighoff2022crosslingual}是一个多语言提示数据集集合，支持46种语言，包含82个数据集。xP3专注于改进模型的跨语言泛化能力。该集合包括从各种语言和文化背景中收集的提示，涵盖了文本生成、信息提取、对话系统等任务。xP3的设计旨在推动多语言模型的发展，使其能够更好地处理和理解不同语言和方言的输入。对训练能够处理多语言任务的大规模语言模型特别有价值，能够为跨语言应用提供重要的支持和资源。

\end{itemize}

\subsection{小结}

本章介绍了预训练阶段后，训练大规模语言模型所使用的微调数据。根据数据的构造方式，本章将现有微调数据分为人工撰写、真实用户数据、基于已有数据集扩充、语言模型直接生成、传统数据集改写五个类别。如表 \ref{sft_summary}所示，这5种微调数据的构建方式有着其各自的特点，每种方法都有其独特的优势和局限性。
在实际应用中，往往会将多种来源的数据结合起来使用，以获得更好的效果。例如，通过人工撰写的高质量指令和回答来作为模型学习回答模式的基础，然后筛选真实用户数据进行补充，这样既可以保证数据的高质量，又可以确保数据贴近真实应用场景和具有更高的多样性。
在使用微调数据进行训练时，往往还需要考虑不同领域数据的配比，补齐基础模型的短板等。
例如，Infinity Instruct数据集包含模型完成该指令所需何种能力与知识的标签，可以在训练时根据能力标签针对性的筛选数据。
总体而言，微调数据的来源可以分为由人类生产和由模型生产。其主要作用在于帮助模型将预训练中获得的语言知识转化为解决具体问题的实际能力。
构建微调数据的目标包括，数据的多样性（能够覆盖各种真实应用的场景），数据质量（答案的准确性、相关性和充分性），安全无害（不包含隐私信息或有害内容）等。构建高质量的微调数据不仅是提升模型实际表现的关键，也是确保模型在现实环境中安全可靠运行的重要保障。

\section{总结}

本论文概述了大语言模型在预训练和微调阶段的数据构建要点。在预训练数据方面，英文数据在数量和质量上优于中文数据，成为训练英文大语言模型的有力资源，而中文数据因网页资源有限、学术数据匮乏、社交媒体平台闭源等因素受限。中文领域开源数据集的缺乏，也阻碍了中文大语言模型的发展，尽管模型可以通过英文数据学习多语言能力，但中文特有的知识和文本风格仍难以完全涵盖。此外，数据隐私与开放性之间的平衡成为关注点，语料枯竭也带来新的挑战，促使研究者探索合成数据等新来源。

在微调数据方面，构建方法多样，包括人工撰写、真实用户数据扩充等。多样性、质量和安全性是微调数据的重要目标，实际应用中常融合多种数据来源以取得最佳效果，并在不同领域数据配比上进行优化。总体而言，预训练与微调数据的高质量构建对于提升大语言模型的性能至关重要。特别在中文领域，各方需协力推动数据来源的拓展与技术改进，使中文大语言模型在各类应用中更具竞争力。

\bibliographystyle{plain}
\bibliography{custom}

\begin{thebibliography}{100}

\bibitem{baai_mtp}
Baai-mtp.
\newblock \url{https://data.baai.ac.cn/details/BAAI-MTP}.

\bibitem{google_bigquery_github}
Bigquery public datasetsy.
\newblock \url{https://cloud.google.com/bigquery/public-data}.

\bibitem{baai_cci}
Cci中文互联网语料库.
\newblock \url{https://data.baai.ac.cn/details/BAAI-CCI}.

\bibitem{codeparrot}
Codeparrot on hugging face.
\newblock \url{https://huggingface.co/codeparrot}.

\bibitem{commoncrawl}
Common crawl.
\newblock \url{https://commoncrawl.org/}.

\bibitem{ibm_project_codenet}
Project codenet.
\newblock \url{https://developer.ibm.com/exchanges/data/all/project-codenet/}.

\bibitem{pubmed_central_pmc}
Pubmed central (pmc).
\newblock \url{https://catalog.data.gov/dataset/pubmed-central-pmc}.

\bibitem{stackoverflow}
Stackoverflow.
\newblock \url{https://stackoverflow.com}.

\bibitem{togetherai}
Together ai.
\newblock \url{https://www.together.ai/}.

\bibitem{BachSYWRNSKBFAD22}
Stephen~H. Bach, Victor Sanh, Zheng~Xin Yong, Albert Webson, Colin Raffel, Nihal~V. Nayak, Abheesht Sharma, Taewoon Kim, M.~Saiful Bari, Thibault F{\'{e}}vry, Zaid Alyafeai, Manan Dey, Andrea Santilli, Zhiqing Sun, Srulik Ben{-}David, Canwen Xu, Gunjan Chhablani, Han Wang, Jason~Alan Fries, Maged~Saeed AlShaibani, Shanya Sharma, Urmish Thakker, Khalid Almubarak, Xiangru Tang, Dragomir~R. Radev, Mike~Tian{-}Jian Jiang, and Alexander~M. Rush.
\newblock Promptsource: An integrated development environment and repository for natural language prompts.
\newblock In {\em {ACL} (demo)}, pages 93--104. Association for Computational Linguistics, 2022.

\bibitem{bai2023qwentechnicalreport}
Jinze Bai, Shuai Bai, Yunfei Chu, Zeyu Cui, Kai Dang, Xiaodong Deng, Yang Fan, Wenbin Ge, Yu~Han, Fei Huang, Binyuan Hui, Luo Ji, Mei Li, Junyang Lin, Runji Lin, Dayiheng Liu, Gao Liu, Chengqiang Lu, Keming Lu, Jianxin Ma, Rui Men, Xingzhang Ren, Xuancheng Ren, Chuanqi Tan, Sinan Tan, Jianhong Tu, Peng Wang, Shijie Wang, Wei Wang, Shengguang Wu, Benfeng Xu, Jin Xu, An~Yang, Hao Yang, Jian Yang, Shusheng Yang, Yang Yao, Bowen Yu, Hongyi Yuan, Zheng Yuan, Jianwei Zhang, Xingxuan Zhang, Yichang Zhang, Zhenru Zhang, Chang Zhou, Jingren Zhou, Xiaohuan Zhou, and Tianhang Zhu.
\newblock Qwen technical report, 2023.

\bibitem{BAAI_infinity_instruct}
{Beijing Academy of Artificial Intelligence (BAAI)}.
\newblock Infinity-instruct dataset, 2023.
\newblock \url{https://huggingface.co/datasets/BAAI/Infinity-Instruct}.

\bibitem{BELLE}
BELLEGroup.
\newblock Belle: Be everyone's large language model engine.
\newblock \url{https://github.com/LianjiaTech/BELLE}, 2023.

\bibitem{bellegroup_generated_chat_0_4M}
{BelleGroup}.
\newblock Generated chat 0.4m dataset, 2023.
\newblock \url{https://huggingface.co/datasets/BelleGroup/generated_chat_0.4M}.

\bibitem{bigcode_self_oss_instruct_sc2_exec_filter_50k}
{bigcode}.
\newblock Self-oss instruct sc2 exec filter 50k dataset, 2023.
\newblock \url{https://huggingface.co/datasets/bigcode/self-oss-instruct-sc2-exec-filter-50k}.

\bibitem{camel_ai_code}
{camel-ai}.
\newblock Code dataset, 2023.
\newblock \url{https://huggingface.co/datasets/camel-ai/code}.

\bibitem{camel_ai_math}
{camel-ai}.
\newblock Math dataset, 2023.
\newblock \url{https://huggingface.co/datasets/camel-ai/math}.

\bibitem{chen2023chinesewebtextlargescalehighqualitychinese}
Jianghao Chen, Pu~Jian, Tengxiao Xi, Dongyi Yi, Qianlong Du, Chenglin Ding, Guibo Zhu, Chengqing Zong, Jinqiao Wang, and Jiajun Zhang.
\newblock Chinesewebtext: Large-scale high-quality chinese web text extracted with effective evaluation model, 2023.

\bibitem{chen2023tigerbotopenmultilingualmultitask}
Ye~Chen, Wei Cai, Liangmin Wu, Xiaowei Li, Zhanxuan Xin, and Cong Fu.
\newblock Tigerbot: An open multilingual multitask llm, 2023.

\bibitem{JosephusCheung_guanaco_dataset}
Joséphus Cheung.
\newblock Guanacodataset, 2023.
\newblock \url{https://huggingface.co/datasets/JosephusCheung/GuanacoDataset}.

\bibitem{cobbe2021training}
Karl Cobbe, Vineet Kosaraju, Mohammad Bavarian, Mark Chen, Heewoo Jun, Lukasz Kaiser, Matthias Plappert, Jerry Tworek, Jacob Hilton, Reiichiro Nakano, et~al.
\newblock Training verifiers to solve math word problems.
\newblock {\em arXiv preprint arXiv:2110.14168}, 2021.

\bibitem{codefuse_ai_codeexercise_python_27k}
{codefuse-ai}.
\newblock Codeexercise-python-27k dataset, 2023.
\newblock \url{https://modelscope.cn/datasets/codefuse-ai/CodeExercise-Python-27k/summary}.

\bibitem{redpajama_data_1t}
Together Computer.
\newblock Redpajama-data-1t dataset on hugging face.
\newblock \url{https://huggingface.co/datasets/togethercomputer/RedPajama-Data-1T}.

\bibitem{bib60}
Mike Conover, Matt Hayes, Ankit Mathur, Jianwei Xie, Jun Wan, Sam Shah, et~al.
\newblock Free {D}olly: Introducing the world's first truly open instruction-tuned {LLM}, 2023.
\newblock \url{https://www.databricks.com/blog/2023/04/12/dolly-first-open-commercially-viable-instruction-tuned-llm}.

\bibitem{ding2023enhancing}
Ning Ding, Yulin Chen, Bokai Xu, Yujia Qin, Zhi Zheng, Shengding Hu, Zhiyuan Liu, Maosong Sun, and Bowen Zhou.
\newblock Enhancing chat language models by scaling high-quality instructional conversations.
\newblock {\em arXiv preprint arXiv:2305.14233}, 2023.

\bibitem{eisele-chen-2010-multiun}
Andreas Eisele and Yu~Chen.
\newblock {M}ulti{UN}: A multilingual corpus from united nation documents.
\newblock In Nicoletta Calzolari, Khalid Choukri, Bente Maegaard, Joseph Mariani, Jan Odijk, Stelios Piperidis, Mike Rosner, and Daniel Tapias, editors, {\em Proceedings of the Seventh International Conference on Language Resources and Evaluation ({LREC}'10)}, Valletta, Malta, May 2010. European Language Resources Association (ELRA).

\bibitem{espla-etal-2019-paracrawl}
Miquel Espl{\`a}, Mikel Forcada, Gema Ram{\'\i}rez-S{\'a}nchez, and Hieu Hoang.
\newblock {P}ara{C}rawl: Web-scale parallel corpora for the languages of the {EU}.
\newblock In Mikel Forcada, Andy Way, John Tinsley, Dimitar Shterionov, Celia Rico, and Federico Gaspari, editors, {\em Proceedings of Machine Translation Summit XVII: Translator, Project and User Tracks}, pages 118--119, Dublin, Ireland, August 2019. European Association for Machine Translation.

\bibitem{flagdata}
FlagOpen.
\newblock Flagdata.
\newblock \url{https://github.com/FlagOpen/FlagData}.

\bibitem{FreedomIntelligence_huatuogpt_sft_data_v1}
{FreedomIntelligence}.
\newblock Huatuogpt-sft data v1, 2023.
\newblock \url{https://huggingface.co/datasets/FreedomIntelligence/HuatuoGPT-sft-data-v1}.

\bibitem{gao2020pile800gbdatasetdiverse}
Leo Gao, Stella Biderman, Sid Black, Laurence Golding, Travis Hoppe, Charles Foster, Jason Phang, Horace He, Anish Thite, Noa Nabeshima, Shawn Presser, and Connor Leahy.
\newblock The pile: An 800gb dataset of diverse text for language modeling, 2020.

\bibitem{garage-bAInd_open_platypus}
{garage-bAInd}.
\newblock Open-platypus dataset, 2023.
\newblock \url{https://huggingface.co/datasets/garage-bAInd/Open-Platypus}.

\bibitem{glaiveai_code_assistant}
{glaiveai}.
\newblock Glaive code assistant dataset, 2023.
\newblock \url{https://huggingface.co/datasets/glaiveai/glaive-code-assistant}.

\bibitem{glm2024chatglmfamilylargelanguage}
Team GLM, :, Aohan Zeng, Bin Xu, Bowen Wang, Chenhui Zhang, Da~Yin, Dan Zhang, Diego Rojas, Guanyu Feng, Hanlin Zhao, Hanyu Lai, Hao Yu, Hongning Wang, Jiadai Sun, Jiajie Zhang, Jiale Cheng, Jiayi Gui, Jie Tang, Jing Zhang, Jingyu Sun, Juanzi Li, Lei Zhao, Lindong Wu, Lucen Zhong, Mingdao Liu, Minlie Huang, Peng Zhang, Qinkai Zheng, Rui Lu, Shuaiqi Duan, Shudan Zhang, Shulin Cao, Shuxun Yang, Weng~Lam Tam, Wenyi Zhao, Xiao Liu, Xiao Xia, Xiaohan Zhang, Xiaotao Gu, Xin Lv, Xinghan Liu, Xinyi Liu, Xinyue Yang, Xixuan Song, Xunkai Zhang, Yifan An, Yifan Xu, Yilin Niu, Yuantao Yang, Yueyan Li, Yushi Bai, Yuxiao Dong, Zehan Qi, Zhaoyu Wang, Zhen Yang, Zhengxiao Du, Zhenyu Hou, and Zihan Wang.
\newblock Chatglm: A family of large language models from glm-130b to glm-4 all tools, 2024.

\bibitem{abs-2305-11527}
Honghao Gui, Shuofei Qiao, Jintian Zhang, Hongbin Ye, Mengshu Sun, Lei Liang, Huajun Chen, and Ningyu Zhang.
\newblock Instructie: {A} bilingual instruction-based information extraction dataset.
\newblock {\em CoRR}, abs/2305.11527, 2023.

\bibitem{abs-2402-14710}
Honghao Gui, Lin Yuan, Hongbin Ye, Ningyu Zhang, Mengshu Sun, Lei Liang, and Huajun Chen.
\newblock Iepile: Unearthing large-scale schema-based information extraction corpus.
\newblock {\em CoRR}, abs/2402.14710, 2024.

\bibitem{guo-etal-2023-hc3}
Biyang Guo, Xin Zhang, Ziyuan Wang, Minqi Jiang, Jinran Nie, Yuxuan Ding, Jianwei Yue, and Yupeng Wu.
\newblock How close is chatgpt to human experts? comparison corpus, evaluation, and detection.
\newblock {\em arXiv preprint arxiv:2301.07597}, 2023.

\bibitem{he2023wanjuancomprehensivemultimodaldataset}
Conghui He, Zhenjiang Jin, Chao Xu, Jiantao Qiu, Bin Wang, Wei Li, Hang Yan, Jiaqi Wang, and Dahua Lin.
\newblock Wanjuan: A comprehensive multimodal dataset for advancing english and chinese large models, 2023.

\bibitem{bib334}
Dan Hendrycks, Collin Burns, Saurav Kadavath, Akul Arora, Steven Basart, Eric Tang, Dawn Song, and Jacob Steinhardt.
\newblock Measuring mathematical problem solving with the {MATH} dataset.
\newblock In {\em Thirty-fifth Conference on Neural Information Processing Systems Datasets and Benchmarks Track (Round 2)}, pages 1--11, 2021.

\bibitem{husain2020codesearchnetchallengeevaluatingstate}
Hamel Husain, Ho-Hsiang Wu, Tiferet Gazit, Miltiadis Allamanis, and Marc Brockschmidt.
\newblock Codesearchnet challenge: Evaluating the state of semantic code search, 2020.

\bibitem{iyer2022opt}
Srinivasan Iyer, Xi~Victoria Lin, Ramakanth Pasunuru, Todor Mihaylov, D{\'a}niel Simig, Ping Yu, Kurt Shuster, Tianlu Wang, Qing Liu, Punit~Singh Koura, et~al.
\newblock Opt-iml: Scaling language model instruction meta learning through the lens of generalization, 2022.

\bibitem{ji2023better}
Yunjie Ji, Yan Gong, Yong Deng, Yiping Peng, Qiang Niu, Baochang Ma, and Xiangang Li.
\newblock Towards better instruction following language models for chinese: Investigating the impact of training data and evaluation, 2023.

\bibitem{jinaai_code_exercises}
{jinaai}.
\newblock Code exercises dataset, 2023.
\newblock \url{https://huggingface.co/datasets/jinaai/code_exercises}.

\bibitem{jtatman_python_code_dataset_500k}
{jtatman}.
\newblock Python code dataset 500k, 2023.
\newblock \url{https://huggingface.co/datasets/jtatman/python-code-dataset-500k}.

\bibitem{kenton2019bert}
Jacob Devlin Ming-Wei~Chang Kenton and Lee~Kristina Toutanova.
\newblock Bert: Pre-training of deep bidirectional transformers for language understanding.
\newblock In {\em Proceedings of NAACL-HLT}, pages 4171--4186, 2019.

\bibitem{bib31}
Denis Kocetkov, Raymond Li, Loubna~Ben allal, Jia LI, Chenghao Mou, Yacine Jernite, Margaret Mitchell, Carlos~Mu{\~n}oz Ferrandis, Sean Hughes, Thomas Wolf, Dzmitry Bahdanau, Leandro~Von Werra, and Harm de~Vries.
\newblock {The Stack}: 3 {TB} of permissively licensed source code.
\newblock {\em Transactions on Machine Learning Research}, pages 1--28, 2023.

\bibitem{bib42}
Hugo Lauren{\c{c}}on, Lucile Saulnier, Thomas Wang, Christopher Akiki, Albert~Villanova del Moral, Teven~Le Scao, Leandro~Von Werra, Chenghao Mou, Eduardo~Gonz{\'a}lez Ponferrada, Huu Nguyen, J{\"o}rg Frohberg, Mario {\v{S}}a{\v{s}}ko, Quentin Lhoest, et~al.
\newblock The {BigScience} {ROOTS} corpus: A 1.6{TB} composite multilingual dataset.
\newblock In {\em Thirty-sixth Conference on Neural Information Processing Systems Datasets and Benchmarks Track}, pages 1--18, 2022.

\bibitem{li2024query}
Chengpeng Li, Zheng Yuan, Hongyi Yuan, Guanting Dong, Keming Lu, Jiancan Wu, Chuanqi Tan, Xiang Wang, and Chang Zhou.
\newblock Query and response augmentation cannot help out-of-domain math reasoning generalization, 2024.

\bibitem{li2023camel}
Guohao Li, Hasan Abed Al~Kader Hammoud, Hani Itani, Dmitrii Khizbullin, and Bernard Ghanem.
\newblock Camel: Communicative agents for "mind" exploration of large scale language model society, 2023.

\bibitem{abs-2305-01526}
Jianquan Li, Xidong Wang, Xiangbo Wu, Zhiyi Zhang, Xiaolong Xu, Jie Fu, Prayag Tiwari, Xiang Wan, and Benyou Wang.
\newblock Huatuo-26m, a large-scale chinese medical {QA} dataset.
\newblock {\em CoRR}, abs/2305.01526, 2023.

\bibitem{abs-2401-09003}
Haoxiong Liu, Yifan Zhang, Yifan Luo, and Andrew~Chi{-}Chih Yao.
\newblock Augmenting math word problems via iterative question composing.
\newblock {\em CoRR}, abs/2401.09003, 2024.

\bibitem{chatbotarena}
LMSYS.
\newblock Chatbot arena: Benchmarking llms in the wild with elo ratings.
\newblock \url{https://lmsys.org}, 2023.

\bibitem{bib43}
Dakuan Lu, Jiaqing Liang, Yipei Xu, Qianyu He, Yipeng Geng, Mengkun Han, Yingsi Xin, Hengkui Wu, and Yanghua Xiao.
\newblock {BBT-Fin}: Comprehensive construction of {C}hinese financial domain pre-trained language model, corpus and benchmark, 2023.
\newblock {a}rXiv preprint \href{https://arxiv.org/abs/2302.09432}{arXiv:2302.09432}.

\bibitem{luo2023wizardcoder}
Ziyang Luo, Can Xu, Pu~Zhao, Qingfeng Sun, Xiubo Geng, Wenxiang Hu, Chongyang Tao, Jing Ma, Qingwei Lin, and Daxin Jiang.
\newblock Wizardcoder: Empowering code large language models with evol-instruct.
\newblock {\em arXiv preprint arXiv:2306.08568}, 2023.

\bibitem{bib26}
{MetaGLM}.
\newblock {FinGLM}, 2023.
\newblock \url{https://github.com/MetaGLM/FinGLM}.

\bibitem{naturalinstructions}
Swaroop Mishra, Daniel Khashabi, Chitta Baral, and Hannaneh Hajishirzi.
\newblock Cross-task generalization via natural language crowdsourcing instructions.
\newblock In {\em ACL}, 2022.

\bibitem{mitra2024orcamath}
Arindam Mitra, Hamed Khanpour, Corby Rosset, and Ahmed Awadallah.
\newblock Orca-math: Unlocking the potential of slms in grade school math, 2024.

\bibitem{bib23}
{MOP-LIWU Community} and {MNBVC Team}.
\newblock {MNBVC}: Massive never-ending {BT} vast {C}hinese corpus, 2023.
\newblock \url{https://github.com/esbatmop/MNBVC}.

\bibitem{muennighoff2022crosslingual}
Niklas Muennighoff, Thomas Wang, Lintang Sutawika, Adam Roberts, Stella Biderman, Teven~Le Scao, M~Saiful Bari, Sheng Shen, Zheng-Xin Yong, Hailey Schoelkopf, et~al.
\newblock Crosslingual generalization through multitask finetuning.
\newblock {\em arXiv preprint arXiv:2211.01786}, 2022.

\bibitem{nan_do_instructional_code_search_net_python}
{Nan-Do}.
\newblock Instructional code-search net python dataset, 2023.
\newblock \url{https://huggingface.co/datasets/Nan-Do/instructional_code-search-net-python}.

\bibitem{bib62}
Jinjie Ni, Fuzhao Xue, Yuntian Deng, Jason Phang, Kabir Jain, Mahir~Hitesh Shah, Zangwei Zheng, and Yang You.
\newblock Instruction in the wild: A user-based instruction dataset, 2023.
\newblock \url{https://github.com/XueFuzhao/InstructionWild}.

\bibitem{ontocord_oig_moderation}
Ontocord.
\newblock Oig-moderation dataset, 2023.
\newblock \url{https://huggingface.co/datasets/ontocord/OIG-moderation}.

\bibitem{openai2022introducing}
OpenAI.
\newblock Introducing chatgpt.
\newblock 2022.
\newblock \url{https://openai.com/research/introducing-chatgpt}.

\bibitem{OpenAssistant_oasst1}
{OpenAssistant}.
\newblock Openassistant oasst1 dataset, 2023.
\newblock \url{https://huggingface.co/datasets/OpenAssistant/oasst1}.

\bibitem{ouyang2022training}
Long Ouyang, Jeffrey Wu, Xu~Jiang, Diogo Almeida, Carroll Wainwright, Pamela Mishkin, Chong Zhang, Sandhini Agarwal, Katarina Slama, Alex Ray, et~al.
\newblock Training language models to follow instructions with human feedback.
\newblock {\em Advances in Neural Information Processing Systems}, 35:27730--27744, 2022.

\bibitem{paster2023openwebmathopendatasethighquality}
Keiran Paster, Marco~Dos Santos, Zhangir Azerbayev, and Jimmy Ba.
\newblock Openwebmath: An open dataset of high-quality mathematical web text, 2023.

\bibitem{penedo2024finewebdatasetsdecantingweb}
Guilherme Penedo, Hynek Kydlíček, Loubna~Ben allal, Anton Lozhkov, Margaret Mitchell, Colin Raffel, Leandro~Von Werra, and Thomas Wolf.
\newblock The fineweb datasets: Decanting the web for the finest text data at scale, 2024.

\bibitem{nips/PenedoMHCACPAL23}
Guilherme Penedo, Quentin Malartic, Daniel Hesslow, Ruxandra Cojocaru, Hamza Alobeidli, Alessandro Cappelli, Baptiste Pannier, Ebtesam Almazrouei, and Julien Launay.
\newblock The refinedweb dataset for falcon {LLM:} outperforming curated corpora with web data only.
\newblock In {\em NeurIPS}, 2023.

\bibitem{peng2023gpt4llm}
Baolin Peng, Chunyuan Li, Pengcheng He, Michel Galley, and Jianfeng Gao.
\newblock Instruction tuning with gpt-4.
\newblock {\em arXiv preprint arXiv:2304.03277}, 2023.

\bibitem{plexpt_chatgpt_corpus}
{PlexPt}.
\newblock Chatgpt corpus, 2023.
\newblock \url{https://github.com/PlexPt/chatgpt-corpus}.

\bibitem{bib52}
Jack~W Rae, Sebastian Borgeaud, Trevor Cai, Katie Millican, Jordan Hoffmann, Francis Song, John Aslanides, Sarah Henderson, Roman Ring, Susannah Young, et~al.
\newblock Scaling language models: Methods, analysis \& insights from training gopher, 2021.
\newblock {a}rXiv preprint \href{https://arxiv.org/abs/2112.11446}{arXiv:2112.11446}.

\bibitem{raffel2020exploring}
Colin Raffel, Noam Shazeer, Adam Roberts, Katherine Lee, Sharan Narang, Michael Matena, Yanqi Zhou, Wei Li, and Peter~J Liu.
\newblock Exploring the limits of transfer learning with a unified text-to-text transformer.
\newblock {\em The Journal of Machine Learning Research}, 21(1):5485--5551, 2020.

\bibitem{RajpurkarJL18}
Pranav Rajpurkar, Robin Jia, and Percy Liang.
\newblock Know what you don't know: Unanswerable questions for squad.
\newblock In {\em {ACL} {(2)}}, pages 784--789. Association for Computational Linguistics, 2018.

\bibitem{RajpurkarZLL16}
Pranav Rajpurkar, Jian Zhang, Konstantin Lopyrev, and Percy Liang.
\newblock Squad: 100, 000+ questions for machine comprehension of text.
\newblock In {\em {EMNLP}}, pages 2383--2392. The Association for Computational Linguistics, 2016.

\bibitem{vicgalle_alpaca_gpt4}
Microsoft Research.
\newblock Alpaca-gpt4 dataset, 2023.
\newblock \url{https://huggingface.co/datasets/vicgalle/alpaca-gpt4}.

\bibitem{sahil2801_codealpaca_20k}
{sahil2801}.
\newblock Codealpaca-20k dataset, 2023.
\newblock \url{https://huggingface.co/datasets/sahil2801/CodeAlpaca-20k}.

\bibitem{sanh2021multitask}
Victor Sanh, Albert Webson, Colin Raffel, Stephen~H. Bach, Lintang Sutawika, Zaid Alyafeai, Antoine Chaffin, Arnaud Stiegler, Teven~Le Scao, Arun Raja, Manan Dey, M~Saiful Bari, Canwen Xu, Urmish Thakker, Shanya~Sharma Sharma, Eliza Szczechla, Taewoon Kim, Gunjan Chhablani, Nihal Nayak, Debajyoti Datta, Jonathan Chang, Mike Tian-Jian Jiang, Han Wang, Matteo Manica, Sheng Shen, Zheng~Xin Yong, Harshit Pandey, Rachel Bawden, Thomas Wang, Trishala Neeraj, Jos Rozen, Abheesht Sharma, Andrea Santilli, Thibault Fevry, Jason~Alan Fries, Ryan Teehan, Stella Biderman, Leo Gao, Tali Bers, Thomas Wolf, and Alexander~M. Rush.
\newblock Multitask prompted training enables zero-shot task generalization, 2021.

\bibitem{ShareGPT-Chinese-English-90k}
shareAI.
\newblock Sharegpt-chinese-english-90k bilingual human-machine qa dataset.
\newblock \url{https://huggingface.co/datasets/shareAI/ShareGPT-Chinese-English-90k}, 2023.

\bibitem{shen2024slimpajamadcunderstandingdatacombinations}
Zhiqiang Shen, Tianhua Tao, Liqun Ma, Willie Neiswanger, Zhengzhong Liu, Hongyi Wang, Bowen Tan, Joel Hestness, Natalia Vassilieva, Daria Soboleva, and Eric Xing.
\newblock Slimpajama-dc: Understanding data combinations for llm training, 2024.

\bibitem{bib409}
Shivalika Singh, Freddie Vargus, Daniel Dsouza, B{\"o}rje~F Karlsson, Abinaya Mahendiran, Wei-Yin Ko, Herumb Shandilya, Jay Patel, Deividas Mataciunas, Laura OMahony, et~al.
\newblock Aya {D}ataset: An open-access collection for multilingual instruction tuning, 2024.
\newblock {a}rXiv preprint \href{https://arxiv.org/abs/2402.06619}{arXiv:2402.06619}.

\bibitem{skypile_150b}
Skywork.
\newblock Skypile-150b dataset.
\newblock \url{https://huggingface.co/datasets/Skywork/SkyPile-150B}.

\bibitem{bib408}
Luca Soldaini, Rodney Kinney, Akshita Bhagia, Dustin Schwenk, David Atkinson, Russell Authur, Ben Bogin, Khyathi Chandu, Jennifer Dumas, Yanai Elazar, Valentin Hofmann, Ananya~Harsh Jha, Sachin Kumar, Li~Lucy, Xinxi Lyu, Ian Magnusson, Jacob Morrison, Niklas Muennighoff, Aakanksha Naik, Crystal Nam, Matthew~E. Peters, Abhilasha Ravichander, Kyle Richardson, Zejiang Shen, Emma Strubell, Nishant Subramani, Oyvind Tafjord, Evan~Pete Walsh, Hannaneh Hajishirzi, Noah~A. Smith, Luke Zettlemoyer, Iz~Beltagy, Dirk Groeneveld, Jesse Dodge, and Kyle Lo.
\newblock Dolma: An open corpus of three trillion tokens for language model pretraining research, 2024.
\newblock {a}rXiv preprint \href{https://browse.arxiv.org/abs/2402.00159}{arXiv:2402.00159}.

\bibitem{SuarezRS20}
Pedro Javier~Ortiz Su{\'{a}}rez, Laurent Romary, and Beno{\^{\i}}t Sagot.
\newblock A monolingual approach to contextualized word embeddings for mid-resource languages.
\newblock In {\em {ACL}}, pages 1703--1714. Association for Computational Linguistics, 2020.

\bibitem{Sun2024MOSS}
Tianxiang Sun, Xiaotian Zhang, Zhengfu He, Peng Li, Qinyuan Cheng, Xiangyang Liu, Hang Yan, Yunfan Shao, Qiong Tang, Shiduo Zhang, Xingjian Zhao, Ke~Chen, Yining Zheng, Zhejian Zhou, Ruixiao Li, Jun Zhan, Yunhua Zhou, Linyang Li, Xiaogui Yang, Lingling Wu, Zhangyue Yin, Xuanjing Huang, Yu-Gang Jiang, and Xipeng Qiu.
\newblock Moss: An open conversational large language model.
\newblock {\em Machine Intelligence Research}, 2024.

\bibitem{taori2023stanford}
Rohan Taori, Ishaan Gulrajani, Tianyi Zhang, Yann Dubois, Xuechen Li, Carlos Guestrin, Percy Liang, and Tatsunori~B Hashimoto.
\newblock Stanford alpaca: An instruction-following llama model, 2023.

\bibitem{OpenHermes2.5}
Teknium.
\newblock Openhermes 2.5: An open dataset of synthetic data for generalist llm assistants, 2023.
\newblock \url{https://huggingface.co/datasets/teknium/OpenHermes-2.5}.

\bibitem{bib16}
Together.
\newblock {RedPajama-Data-v2}: An open dataset with 30 trillion tokens for training large language models, 2023.
\newblock \url{https://www.together.ai/blog/redpajama-data-v2}.

\bibitem{abs-2407-13690}
Yuxuan Tong, Xiwen Zhang, Rui Wang, Ruidong Wu, and Junxian He.
\newblock Dart-math: Difficulty-aware rejection tuning for mathematical problem-solving.
\newblock {\em CoRR}, abs/2407.13690, 2024.

\bibitem{bib411}
Shubham Toshniwal, Ivan Moshkov, Sean Narenthiran, Daria Gitman, Fei Jia, and Igor Gitman.
\newblock {OpenMathInstruct-1}: A 1.8 million math instruction tuning dataset, 2024.
\newblock {a}rXiv preprint \href{https://arxiv.org/abs/2402.10176}{arXiv:2402.10176}.

\bibitem{touvron2023llama}
Hugo Touvron, Louis Martin, Kevin Stone, Peter Albert, Amjad Almahairi, Yasmine Babaei, Nikolay Bashlykov, Soumya Batra, Prajjwal Bhargava, Shruti Bhosale, et~al.
\newblock Llama 2: Open foundation and fine-tuned chat models.
\newblock {\em arXiv preprint arXiv:2307.09288}, 2023.

\bibitem{BillGPT_chinese_medical_dialogue}
{Toyhom}.
\newblock Chinese medical dialogue data, 2023.
\newblock \url{https://huggingface.co/datasets/BillGPT/Chinese-medical-dialogue-data}.

\bibitem{vaswani2017attention}
Ashish Vaswani, Noam Shazeer, Niki Parmar, Jakob Uszkoreit, Llion Jones, Aidan~N Gomez, {\L}ukasz Kaiser, and Illia Polosukhin.
\newblock Attention is all you need.
\newblock {\em Advances in neural information processing systems}, 30, 2017.

\bibitem{WangCZLSL24}
Guan Wang, Sijie Cheng, Xianyuan Zhan, Xiangang Li, Sen Song, and Yang Liu.
\newblock Openchat: Advancing open-source language models with mixed-quality data.
\newblock In {\em {ICLR}}. OpenReview.net, 2024.

\bibitem{wangrui6_zhihu_kol}
Rui Wang.
\newblock Zhihu-kol dataset, 2023.
\newblock \url{https://huggingface.co/datasets/wangrui6/Zhihu-KOL}.

\bibitem{WangKZHJZH20}
Yida Wang, Pei Ke, Yinhe Zheng, Kaili Huang, Yong Jiang, Xiaoyan Zhu, and Minlie Huang.
\newblock A large-scale chinese short-text conversation dataset.
\newblock In {\em {NLPCC} {(1)}}, volume 12430 of {\em Lecture Notes in Computer Science}, pages 91--103. Springer, 2020.

\bibitem{WangKMLSKH23}
Yizhong Wang, Yeganeh Kordi, Swaroop Mishra, Alisa Liu, Noah~A. Smith, Daniel Khashabi, and Hannaneh Hajishirzi.
\newblock Self-instruct: Aligning language models with self-generated instructions.
\newblock In {\em {ACL} {(1)}}, pages 13484--13508. Association for Computational Linguistics, 2023.

\bibitem{supernaturalinstructions}
Yizhong Wang, Swaroop Mishra, Pegah Alipoormolabashi, Yeganeh Kordi, Amirreza Mirzaei, Anjana Arunkumar, Arjun Ashok, Arut~Selvan Dhanasekaran, Atharva Naik, David Stap, et~al.
\newblock Super-naturalinstructions:generalization via declarative instructions on 1600+ tasks.
\newblock In {\em EMNLP}, 2022.

\bibitem{WangMAKMNADASPK22}
Yizhong Wang, Swaroop Mishra, Pegah Alipoormolabashi, Yeganeh Kordi, Amirreza Mirzaei, Atharva Naik, Arjun Ashok, Arut~Selvan Dhanasekaran, Anjana Arunkumar, David Stap, Eshaan Pathak, Giannis Karamanolakis, Haizhi~Gary Lai, Ishan Purohit, Ishani Mondal, Jacob Anderson, Kirby Kuznia, Krima Doshi, Kuntal~Kumar Pal, Maitreya Patel, Mehrad Moradshahi, Mihir Parmar, Mirali Purohit, Neeraj Varshney, Phani~Rohitha Kaza, Pulkit Verma, Ravsehaj~Singh Puri, Rushang Karia, Savan Doshi, Shailaja~Keyur Sampat, Siddhartha Mishra, Sujan~Reddy A, Sumanta Patro, Tanay Dixit, and Xudong Shen.
\newblock Super-naturalinstructions: Generalization via declarative instructions on 1600+ {NLP} tasks.
\newblock In {\em {EMNLP}}, pages 5085--5109. Association for Computational Linguistics, 2022.

\bibitem{iclr/WeiBZGYLDDL22}
Jason Wei, Maarten Bosma, Vincent~Y. Zhao, Kelvin Guu, Adams~Wei Yu, Brian Lester, Nan Du, Andrew~M. Dai, and Quoc~V. Le.
\newblock Finetuned language models are zero-shot learners.
\newblock In {\em {ICLR}}. OpenReview.net, 2022.

\bibitem{wenzek-etal-2020-ccnet}
Guillaume Wenzek, Marie-Anne Lachaux, Alexis Conneau, Vishrav Chaudhary, Francisco Guzm{\'a}n, Armand Joulin, and Edouard Grave.
\newblock {CCN}et: Extracting high quality monolingual datasets from web crawl data.
\newblock In Nicoletta Calzolari, Fr{\'e}d{\'e}ric B{\'e}chet, Philippe Blache, Khalid Choukri, Christopher Cieri, Thierry Declerck, Sara Goggi, Hitoshi Isahara, Bente Maegaard, Joseph Mariani, H{\'e}l{\`e}ne Mazo, Asuncion Moreno, Jan Odijk, and Stelios Piperidis, editors, {\em Proceedings of the Twelfth Language Resources and Evaluation Conference}, pages 4003--4012, Marseille, France, May 2020. European Language Resources Association.

\bibitem{bib68}
Can Xu, Qingfeng Sun, Kai Zheng, Xiubo Geng, Pu~Zhao, Jiazhan Feng, Chongyang Tao, and Daxin Jiang.
\newblock Wizard{LM}: Empowering large language models to follow complex instructions, 2023.
\newblock {a}rXiv preprint \href{https://arxiv.org/abs/2304.12244}{arXiv:2304.12244}.

\bibitem{bib20}
Liang Xu, Xuanwei Zhang, and Qianqian Dong.
\newblock {CLUECorpus2020}: A large-scale {C}hinese corpus for pre-training language model, 2020.
\newblock {a}rXiv preprint \href{https://arxiv.org/abs/2003.01355}{arXiv:2003.01355}.

\bibitem{bib45}
Ming Xu.
\newblock {MedicalGPT}: Training medical {GPT} model, 2023.
\newblock \url{https://github.com/shibing624/MedicalGPT}.

\bibitem{bib5}
Aiyuan Yang, Bin Xiao, Bingning Wang, Borong Zhang, Ce~Bian, Chao Yin, Chenxu Lv, Da~Pan, Dian Wang, Dong Yan, et~al.
\newblock Baichuan 2: Open large-scale language models, 2023.
\newblock {a}rXiv preprint \href{https://arxiv.org/abs/2309.10305}{arXiv:2309.10305}.

\bibitem{redgpt}
Dongjie Yang, Ruifeng Yuan, Yuantao Fan, , Yifei Yang, Zili Wang, and Shusen Wang.
\newblock Redgpt: Reference-enlightened-dialogue by gpt and for gpt.
\newblock \url{https://github.com/ziliwangnlp/RedGPT}, 2023.

\bibitem{ye-etal-2021-crossfit}
Qinyuan Ye, Bill~Yuchen Lin, and Xiang Ren.
\newblock {C}ross{F}it: A few-shot learning challenge for cross-task generalization in {NLP}.
\newblock pages 7163--7189, November 2021.

\bibitem{yeungnlp_firefly_train_1_1M}
{YeungNLP}.
\newblock Firefly train 1.1m dataset, 2023.
\newblock \url{https://huggingface.co/datasets/YeungNLP/firefly-train-1.1M}.

\bibitem{bib76}
Da~Yin, Xiao Liu, Fan Yin, Ming Zhong, Hritik Bansal, Jiawei Han, and Kai-Wei Chang.
\newblock Dynosaur: A dynamic growth paradigm for instruction-tuning data curation.
\newblock In {\em The 2023 Conference on Empirical Methods in Natural Language Processing}, pages 1--17, 2023.

\bibitem{yu2023metamath}
Longhui Yu, Weisen Jiang, Han Shi, Jincheng Yu, Zhengying Liu, Yu~Zhang, James~T Kwok, Zhenguo Li, Adrian Weller, and Weiyang Liu.
\newblock Metamath: Bootstrap your own mathematical questions for large language models.
\newblock {\em arXiv preprint arXiv:2309.12284}, 2023.

\bibitem{bib22}
Sha Yuan, Hanyu Zhao, Zhengxiao Du, Ming Ding, Xiao Liu, Yukuo Cen, Xu~Zou, Zhilin Yang, and Jie Tang.
\newblock {WuDaoCorpora}: A super large-scale {C}hinese corpora for pre-training language models.
\newblock {\em AI Open}, 2:65--68, 2021.

\bibitem{ZhangQLHMLYWSX24}
Jianguo Zhang, Kun Qian, Zhiwei Liu, Shelby Heinecke, Rui Meng, Ye~Liu, Zhou Yu, Huan Wang, Silvio Savarese, and Caiming Xiong.
\newblock Dialogstudio: Towards richest and most diverse unified dataset collection for conversational {AI}.
\newblock In {\em {EACL} (Findings)}, pages 2299--2315. Association for Computational Linguistics, 2024.

\bibitem{bib44}
Xuanyu Zhang and Qing Yang.
\newblock {XuanYuan 2.0}: A large {C}hinese financial chat model with hundreds of billions parameters.
\newblock In {\em Proceedings of the 32nd ACM International Conference on Information and Knowledge Management}, pages 4435--4439, New York, NY, USA, 2023. ACM.

\bibitem{Zhao0HC0D24}
Wenting Zhao, Xiang Ren, Jack Hessel, Claire Cardie, Yejin Choi, and Yuntian Deng.
\newblock Wildchat: 1m chatgpt interaction logs in the wild.
\newblock In {\em {ICLR}}. OpenReview.net, 2024.

\bibitem{zheng2023lmsys}
Lianmin Zheng, Wei-Lin Chiang, Ying Sheng, Tianle Li, Siyuan Zhuang, Zhanghao Wu, Yonghao Zhuang, Zhuohan Li, Zi~Lin, Eric Xing, et~al.
\newblock Lmsys-chat-1m: A large-scale real-world llm conversation dataset.
\newblock {\em arXiv preprint arXiv:2309.11998}, 2023.

\bibitem{ZhengZSLLFCY24}
Tianyu Zheng, Ge~Zhang, Tianhao Shen, Xueling Liu, Bill~Yuchen Lin, Jie Fu, Wenhu Chen, and Xiang Yue.
\newblock Opencodeinterpreter: Integrating code generation with execution and refinement.
\newblock In {\em {ACL} (Findings)}, pages 12834--12859. Association for Computational Linguistics, 2024.

\bibitem{bib90}
Chunting Zhou, Pengfei Liu, Puxin Xu, Srini Iyer, Jiao Sun, Yuning Mao, Xuezhe Ma, Avia Efrat, Ping Yu, Lili Yu, et~al.
\newblock {LIMA}: Less is more for alignment, 2023.
\newblock {a}rXiv preprint \href{https://arxiv.org/abs/2305.11206}{arXiv:2305.11206}.

\bibitem{ziemski-etal-2016-united}
Micha{\l} Ziemski, Marcin Junczys-Dowmunt, and Bruno Pouliquen.
\newblock The {U}nited {N}ations parallel corpus v1.0.
\newblock In Nicoletta Calzolari, Khalid Choukri, Thierry Declerck, Sara Goggi, Marko Grobelnik, Bente Maegaard, Joseph Mariani, Helene Mazo, Asuncion Moreno, Jan Odijk, and Stelios Piperidis, editors, {\em Proceedings of the Tenth International Conference on Language Resources and Evaluation ({LREC}'16)}, pages 3530--3534, Portoro{\v{z}}, Slovenia, May 2016. European Language Resources Association (ELRA).

\end{thebibliography}

\end{CJK*}
\end{document}